%% file: main.tex
\pdfoutput=1

\documentclass[11pt,
a4paper,
pagebackref
]{article}
\usepackage{siunitx}
\usepackage[final]{acl}
\definecolor{pastelblue}{HTML}{6E8CFB}
\definecolor{goodred}{HTML}{E45A92}

\hypersetup{
    backref=true,
    colorlinks=true,
    linkcolor=pastelblue, %12,13 숫자들
    citecolor=pastelblue, % 인용되는 글씨
    urlcolor=black,
    breaklinks
}
\usepackage{natbib}
\usepackage[most]{tcolorbox}
\usepackage{times}
\usepackage{latexsym}
\usepackage{kotex}
\usepackage{tikz}
\usepackage{enumitem}
\usepackage{siunitx}
\usepackage{adjustbox}
\usepackage{makecell}
\usepackage{colortbl} 
\usepackage{float}

\usetikzlibrary{arrows.meta,shapes,positioning,shadows,fit,backgrounds}
\usepackage[utf8]{inputenc}

\usepackage{inconsolata}

\usepackage{graphicx}
\usepackage{tcolorbox}
\usepackage{listings}
\tcbuselibrary{listingsutf8}

\usepackage{booktabs}
\usepackage{multirow}
\usepackage{pifont}  % for \ding
\usepackage{xcolor}

\title{\textsc{Talk to Your Slides}: High-Efficiency Slide Editing via Language-Driven Structured Data Manipulation}
\author{
 \textbf{Kyudan Jung\textsuperscript{1}},
 \textbf{Hojun Cho\textsuperscript{1}},
 \textbf{Jooyeol Yun\textsuperscript{1}},
 \textbf{Soyoung Yang\textsuperscript{1}},
 \textbf{Jaehyeok Jang\textsuperscript{2}},
 \textbf{Jaegul Choo\textsuperscript{1}},
\\
   \text{$^1$KAIST AI, $^2$Chung-Ang University}\\
   \texttt{\{kyudan, hojun.cho, blizzard072, sy\_yang\}@kaist.ac.kr, achilloaaa@cau.ac.kr} \\
   \texttt{jchoo@kaist.ac.kr}
}

\begin{document}

\maketitle

\begin{abstract}
Editing presentation slides is a frequent yet tedious task, ranging from creative layout design to repetitive text maintenance.
While recent GUI-based agents powered by Multimodal LLMs (MLLMs) excel at tasks requiring visual perception, such as spatial layout adjustments, they often incur high computational costs and latency when handling structured, text-centric, or batch processing tasks.
In this paper, we propose \textsc{Talk-to-Your-Slides}, a high-efficiency slide editing agent that operates via language-driven structured data manipulation rather than relying on the image modality.
By leveraging the underlying object model instead of screen pixels, our approach ensures precise content modification while preserving style fidelity, addressing the limitations of OCR-based visual agents.
Our system features a hierarchical architecture that effectively bridges high-level user instructions with low-level execution codes.
Experiments demonstrate that for text-centric and formatting tasks, our method enables 34\% faster processing, achieves 34\% better instruction fidelity, and operates at an 87\% lower cost compared to GUI-based baselines.
Furthermore, we introduce \textsc{TSBench}, a human-verified benchmark dataset comprising 379 instructions, including a \textsc{Hard} subset designed to evaluate robustness against complex and visually dependent queries.
Our code and benchmark are available at \href{https://github.com/KyuDan1/Talk-to-Your-Slides}{\textbf{\color{pastelblue}{here}}}.

% https://github.com/KyuDan1/Talk-to-Your-Slides
% \href{https://github.com/KyuDan1/Talk-to-Your-Slides}{\textbf{\color{pastelblue}{https://github.com/KyuDan1/Talk-to-Your-Slides}}}.

\end{abstract}

%Fig 1에 대한 업데이트 -introduction글 해야됨.

% 다시 쓰는 Intro.
\begin{figure}[t]
\centering
\includegraphics[width=1\linewidth, trim={250 180 250 180}, clip]{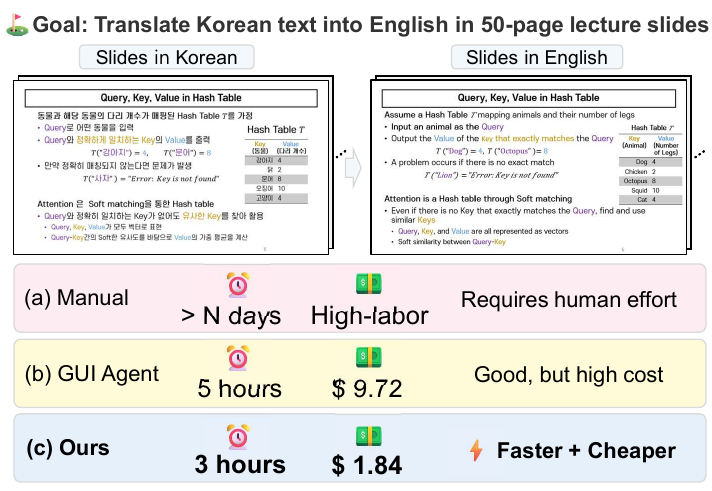}
\caption{
Comparison of slide editing methods on translating 50-page lecture slides from Korean to English. (a) Manual translation requires day(s) and consumes graduate-student labor. (b) A GUI-based agent incurs high cost. (c) Our approach runs in a low cost and in a relatively short time.}
\label{fig:1}
\end{figure}
\section{Introduction}
\label{introduction}

\begin{figure*}[t]
    \centering
    \includegraphics[width=1\linewidth, trim={75 220 85 200}, clip]{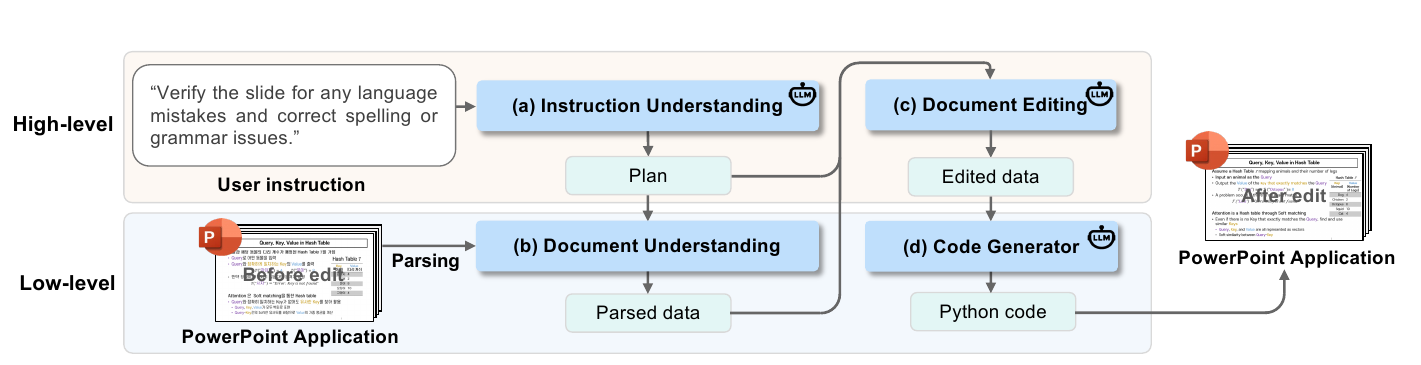}
    \caption{Overview of the \textsc{Talk-to-Your-Slides} framework. The system consists of four modules: instruction understanding, document understanding, document editing, and code generator.}
    \label{fig:overview}
\end{figure*}

Recent advancements in large language models (LLMs) have revolutionized software automation, demonstrating remarkable success in code generation~\citep{yang2024if,hou2024largelanguagemodelssoftware,xu2025skilldiscoverysoftwarescripting}, GUI navigation~\citep{zhang2024ufo, hong2024cogagent}, and slide generation~\citep{sefid2021slidegen, zheng2025pptagentgeneratingevaluatingpresentations}.
However, slide tasks are often bifurcated into \textit{creative design} and \textit{content maintenance}.
While automated generation has garnered attention, there is a growing demand for the latter, efficiently editing existing slides to reflect updated information or formatting.
Slides serve as a fundamental medium for communication, yet modifying them often demands tedious, time-consuming manual effort, particularly for text-centric batch processing.

For example, as shown in Figure~\ref{fig:1}, a professor prepares an international lecture containing 50 slides and must translate them from Korean to English while strictly preserving technical terminology and formatting.
Similarly, a marketing team needs to update product pricing on 120 slides spanning several presentations before a major launch.
In these scenarios, manual effort is prone to errors and inefficiency.

Several candidate approaches can address these challenges.
One straightforward approach is leveraging vision-based GUI agents that operate on screen-captured images through mouse and keyboard interactions~\citep{zhang2024ufo,Microsoft365Copilot}.
While these agents excel at spatial tasks, applying them to text-heavy or batch-editing tasks reveals significant limitations due to the high computational cost of processing image inputs and the potential loss of fidelity in text recognition.
This leads to our first research question \textbf{(RQ1)}: \textit{\textbf{Can a non-visual, structure-aware agent achieve superior efficiency and fidelity for text-centric slide editing compared to visual approaches?}}
We acknowledge that PowerPoint is an inherently visual medium; a purely non-visual agent cannot fully address layout stability (e.g., text overflow after translation) or subjective aesthetic judgments. Our scope is therefore deliberately focused on \textit{structured, text-centric, and batch processing tasks}, where language-driven manipulation offers clear efficiency gains.

Another approach involves converting natural language instructions into direct scripting code.
However, simple baselines struggle with complex tasks requiring contextual interpretation.
For instance, instructions such as ``Summarize the text on all slides and highlight key points in red'' demand understanding both user intent and the sequential context of slide content.
We therefore pose our second research question \textbf{(RQ2)}: \textit{\textbf{What system architecture enables the effective decomposition of complex instructions into precise executable steps?}}

In this paper, we answer our two research questions and propose \textsc{Talk-to-Your-Slides}, an LLM-powered agent designed for efficient, high-fidelity slide editing.
Unlike generalist visual agents, our method operates by leveraging the underlying structured data (e.g., XML, VBA object) of slide objects.
This approach enables more accurate and cost-effective editing by avoiding the overhead of image processing.
Experiments demonstrate that our method is significantly more cost-effective than UFO~\citep{zhang2024ufo}, the current state-of-the-art open-source UI agent.

We design the agent architecture with distinct high-level and low-level layers to facilitate interaction between user commands and slide objects.
This hierarchical design ensures that complex logical instructions are accurately translated into executable code, drawing from insights on planning and reasoning processes~\citep{guo-etal-2024-controllable,chen-etal-2024-towards-tool}.
By providing direct access to application objects, we achieve faster processing and better instruction fidelity compared to GUI-based methods.

Moreover, to complement benchmarks focused on visual aesthetics~\citep{ge2025autopresentdesigningstructuredvisuals, zheng2025pptagentgeneratingevaluatingpresentations}, we present \textit{TSBench}, a human-verified benchmark designed to evaluate slide editing capabilities.
TSBench consists of 379 diverse editing instructions, covering text editing, visual formatting, layout adjustment, and structure manipulation.
Crucially, to address task complexity, we include a robust subset, TSBench-Hard, which challenges the agent with visually dependent or ambiguous tasks, enabling a systematic assessment of the agent's reasoning capabilities.

Our contributions are summarized as follows:
\begin{itemize}
\item We demonstrate that for \textbf{text-centric and batch editing tasks}, a language-driven structural agent significantly outperforms GUI-based agents in speed, cost-efficiency, and instruction fidelity.

\item We introduce \textsc{Talk-to-Your-Slides}, a hierarchical framework specifically designed to bridge natural language instructions with low-level slide objects.

\item We construct \textit{TSBench} (including the challenging \textit{Hard} subset), a benchmark that enables systematic evaluation of slide editing agents beyond simple generation tasks.
\end{itemize}

\begin{figure*}[t]
    \centering
    \includegraphics[width=0.95\textwidth, trim={85 145 85 145}, clip]{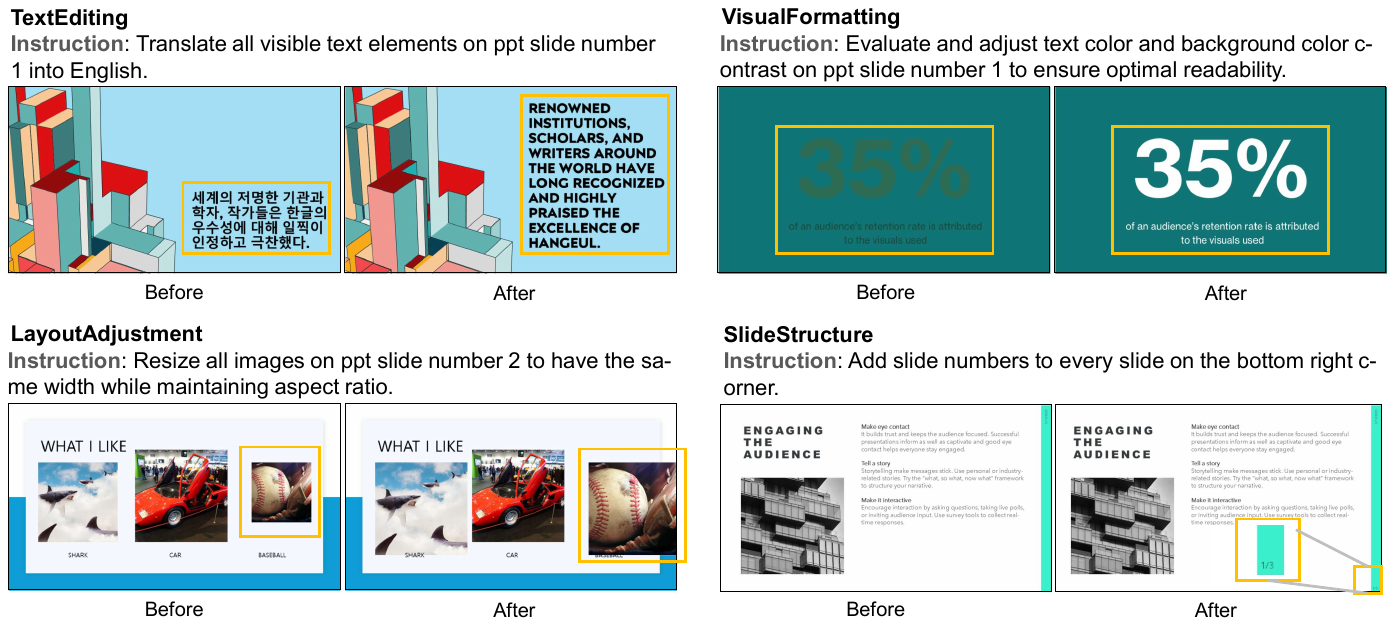}
    \caption{Results of \textsc{Talk-to-Your-Slides} across four instruction categories. Modifications are highlighted with yellow boxes. \textit{TextEditing}: Korean text has been translated into English according to the instruction. \textit{VisualFormatting}: the original background and text colors were too similar, reducing readability; the revised version uses white text for improved contrast and clarity. \textit{LayoutAdjustment}: the widths of the three images have been unified while preserving their aspect ratios, as instructed. \textit{SlideStructure}: the page numbers have been added in response to the instruction.}
\label{example_fig}
\end{figure*}

\section{Related Work}
LLM agents are garnering significant interest in the modern era for their potential to perform complex tasks on behalf of humans. This section specifically reviews related research in three areas: slide generation, GUI-based LLM agents, and code generation.

\subsection{Slide Generation}
\label{related:slide-generation}
Prior work has primarily focused on generating presentation slides from natural language descriptions~\citep{sefid2021slidegen}.
AutoPresent~\citep{ge2025autopresentdesigningstructuredvisuals} fine-tuned an LLaMA-based model on the SlidesBench training set, a dataset comprising 7k slide-generation examples, to generate python code that invokes the SlidesLib API. This approach remains prone to execution errors and does not support fine-grained slide editing.
PPTAgent~\citep{zheng2025pptagentgeneratingevaluatingpresentations} presents a simple process that mimics how people author slides. It first creates an outline and then edits slides using a fixed template. It also includes PPTEval, a tool to check slide content, design, and structure. PPTAgent works well for generating new slides. Our research extends these approaches by introducing editing existing slides capabilities that significantly reduce the manual effort required from users.
\subsection{LLM Agents for GUI Control}
LLM-based agents that control graphical user interfaces (GUIs) \citep{assistgui2024} and \citep{koh2024visualwebarena} are also an active area of research. UFO and UFO2 by Microsoft \citep{zhang2024ufo} introduces a dual-agent framework composed of an application-selection agent and an action agent that can operate across Windows applications such as Word and PowerPoint. By observing application screenshots, the agent executes actions like menu clicks and text input. While powerful, UFO relies on image-based state representations and pixel-level interactions, which can introduce high computational costs and imprecise behavior, particularly for complex editing tasks. We compare our system against this model as a baseline.

\subsection{Code Generation from language instructions}
The task of translating natural language instructions into executable code has attracted considerable attention with the emergence of LLMs. While early work \citet{NL2CodeASurvey} relied on rule-based systems or domain-specific languages-approaches that often lacked scalability and adaptability, LLMs have enabled more flexible, generalizable solutions~\citep{jiang2024surveylargelanguagemodels, yin-etal-NL-to-code-gen}.

Building on this progress, recent studies have introduced intermediate reasoning steps to further enhance code generation, such as guiding code generation through explicit natural language planning~\citep{wang2025planning}, using intermediate plans to decompose and solve complex~\citep{sun-etal-2024-enhancing-code}, multi-step coding tasks-thereby bridging the gap between high-level user intent and low-level executable code~\citep{puerto-etal-2024-code,yang2025codethinkthinkcode}. We utilize this code generation idea in our system to translate user instructions into slide editing operations.

%%%%%%%%%%%%%%%%%%%%%%%%%%%%%%%%%%%%%
%%%%%%%%%% Proposed Method %%%%%%%%%%
%%%%%%%%%%%%%%%%%%%%%%%%%%%%%%%%%%%%%

\section{Method}
\label{method}
% To edit PowerPoint slides according to a user's instruction, four key capabilities are required.

For clarity, we first categorize the capabilities required to edit slides based on a user's instruction into four key components.

First, the system must accurately understand the user's instruction.
Second, to implement this instruction, it needs to comprehend the current state of the presentation slides.
Third, based on the instruction and current state, it should generate slide data that reflects the instruction.
Finally, it must implement these generated changes in the presentation environment such as PowerPoint.

These sequential requirements naturally divide slide editing tasks into two levels. High level operations involve instruction interpretation and content editing. Low level operations require direct access to and manipulation of the presentation software ~\citep{caldiran2009bridging}.

With these careful consideration, we propose \textsc{Talk-to-Your-Slides}, a system that separates these concerns into high-level and low-level components, using language modality as illustrated in Figure~\ref{fig:overview}. In the remainder of this section, we describe each components in detail.
\input{main_table_edit}
\subsection{High-level: Instruction understanding}
\label{Planner}
For an LLM agent to edit slides, the first and most critical step is to understand the user's instructions.
To this end, we follow the planning methodology described in \citet{ruan2023tptu}. Our instruction understanding module operates at a high level within our system architecture, interpreting user instructions into structured, actionable plans that specify which slides to modify, elements to target, and actions to perform, as shown in Figure~\ref{fig:overview}(a). The module outputs a structured list where each entry explicitly details the target slide number, the targeted element, and the corresponding action, allowing for precise and versatile slide editing tasks as shown in Figure~\ref{fig:planner}.

To handle diverse instructions that may target specific slides, subsets, or entire presentations, we utilize an LLM with carefully crafted prompts to support effective in-context learning as illustrated in~\citet{dong-etal-2024-survey} and \citet{guo2024how}. The concrete prompts used for instruction understanding is included in Appendix~\ref{planner_prompt}. Also an example of this module's output and the corresponding original slide can be found in Figure~\ref{fig:planner} and Figure~\ref{fig:parser-image}, respectively.

%ppt 그림도 같이 제시해야될 것 같음.

\subsection{Low-level: Document understanding}
\label{Parser}
After understanding the user's command, a comprehensive understanding of the slide document is essential. Document understanding is a low-level component that accesses slides in our system (see Figure~\ref{fig:overview}). It plays a crucial role in slide editing tasks, as the quality of the parsed content essentially defines the initial set of editable elements and thereby determines an upper bound on the final editing accuracy. 

To enable fine-grained document understanding, we developed a custom rule-based parser that extracts comprehensive information from each slide. This includes metadata such as the layout name, background fill type, and transition effects, as well as fine-level attributes of individual objects such as shapes, images, and text boxes. As shown in Figure~\ref{fig:parser}, the parser identifies both the semantic type and positional information of each element on the slide. The parsed original slide that produced the results in Figure~\ref{fig:parser} is shown in Figure~\ref{fig:parser-image}.

In slide editing programs, a single text placeholder can contain text with diverse font formatting. Therefore, we found that we must parse the text at the \textit{run-level} rather than at the placeholder level, where each \textit{run} denotes a contiguous segment of text with consistent formatting. This level of details allow the system to more faithfully reflect how humans perceive and manipulate slides, thereby enabling precise, style-preserving edits.

We followed the finding that structured data formats enable LLMs to produce better outputs~\citep{he2024doespromptformattingimpact, Struct-X}. Accordingly, parsed outputs are converted into a structured JSON format. This representation facilitates downstream reasoning and editing, while ensuring compatibility with large language models. Our parsing logic is fully reproducible to support future research in structured document understanding. Additional details on document understanding are provided in Appendix~\ref{app:doc-understanding}.

\subsection{High-level: Document editing}
\label{Processor}
With an understanding of both the user's instruction and the document's content, it is time to edit the slide document. The role of the document editing module is to modify the parsed content using an LLM, in accordance with the editing plan generated by the two preceding modules.
For example, if the requirement specifies changing only the important content in a text box to red, the document editing module identifies such content from the parsed data and generates output in the same format as that produced by the document understanding module. The prompt used for this module is provided in Appendix~\ref{applier_prompt}.

\subsection{Low-level: Code generator}
\label{Applier}
% code generator 기능
The primary function of code generator is to generate python code that applies the necessary modifications to the slide editing program instance.
% input output
It receives the raw parsed data before edited, the data after edited from document editing module, and the plan as illustrated in Figure~\ref{fig:overview}.
Once this module generates the code, it is then applied to the program. For the Windows version of PowerPoint, we adopt the COM protocol~\citep{Microsoft2021COM} with details provided in Appendix~\ref{app:COM}. For macOS, AppleScript~\citep{AppleScriptGuide} can be used, as described in Appendix~\ref{app:AppleScript}. Additionally, an MCP server~\citep{AnthropicMCP2025} is another alternative, which we discuss in Appendix~\ref{app:MCP}.
%
% Consequently, the LLM is tasked with generating python code based on the semantics of presentation environment  \footnote{E.g, VBA API for Powerpoint: https://learn.microsoft.com/en-us/office/vba/api/overview/powerpoint}.

% 이러한 실제 슬라이드를 수정하는 부분을 low level로 분리하는 것은 두가지 이점이 있다.
% 첫번째로 Powerpoint와 같은 presentation environment에 대한 dependency를 최소화할 수 있고, 이 때문에 high level 수정방향 생성과정에서 LLM의 부담을 줄이는 이점이 있다.
% 두번째는 Powerpoint아닌 다른 application으로 교체하는 경우, high level modification은 영향을 끼치지 않고, 본 component만 바꾸면 되기에 정확도 변화를 최소화 할 수 있다.

% Separating the part that modifies actual slides at a low-level has a significant advantage.
% This modular design enhances platform adaptability. If the presentation application changes from PowerPoint to another platform, only the low-level component needs modification. The high-level reasoning remains intact. This minimizes accuracy degradation when transitioning between different presentation software.

In addition, we follow the self-reflection mechanism described in \citep{shinn2023reflexion}, to handle execution failures. We show the real world example on Appendix~\ref{app:self-relfection-example}. If an error occurs during execution, the error message and the generated code are appended to the original input, and inference is repeated until the modification succeeds or a predefined maximum number of iterations is reached. The prompt used for the code generator is described in Appendix~\ref{applier_prompt}.

%%%%%%%%%%%%%%%%%%%%%%%%%%
%%%%%%% Benchmark %%%%%%%%

\section{TSBench: Benchmark Dataset}
Alongside proposing a novel system for editing presentation slides, we identify a notable scarcity of evaluation benchmarks for realistic, user-centric slide editing commands ~\citep{guo2023pptcbenchmarkevaluatinglarge, zhang2024pptcrbenchmarkevaluatingrobustness, zheng2025pptagentgeneratingevaluatingpresentations}. To address this gap and facilitate modular testing across a wide range of practical commands, we introduce TSBench, a new benchmark dataset designed to evaluate the slide editing capabilities of models or frameworks. In this section, we describe the construction process of our benchmark dataset in detail and present its key statistics.

\input{human_eval_performance_gemini}

\subsection{Building the Benchmark}
Users often perform edits on large presentations, with a single command sometimes impacting over 50 slides. However, rather than preparing such large-scale documents for evaluation, our benchmark is designed to deconstruct these complex scenarios into fundamental, modular tasks. This approach allows for a granular assessment of an agent’s core capabilities without the overhead of managing massive slide decks.

To build this benchmark, we first created a dataset of user instructions and then developed the corresponding slide data to which these instructions could be applied. In this section, we describe the construction methodology for each component in detail.

\paragraph{Instructions data}To collect feasible and practical instructions, we first manually create 56 seed instructions that reflect plausible user commands. For each seed, we used GPT-4o~\citep{openai2024gpt4o} to generate 10 variations or paraphrases. These variations include, for example, replacing the target language in ``Translate the slide into Chinese'' with alternatives such as Japanese, French, or English. Some variations also involve paraphrasing while preserving the original intent. From the generated pool, we manually filtered out instructions lacking clear goals or evaluation criteria. Given that the precise interpretation of instructions significantly impacts evaluation outcomes, we implemented a manual review process where human evaluators examined all instructions to ensure clarity and validity. Only those with unambiguous objectives were retained.

In total, we collected 379 instructions, which are categorized into four types: \textit{TextEditing}, \textit{VisualFormatting}, \textit{LayoutAdjustment}, and \textit{SlideStructure}. These categories are also used as evaluation dimensions in our analysis. Examples of instructions for each category are illustrated in Figure~\ref{inst_example}.

%instruction 예시 category 별.
\paragraph{Slide deck data}Constructing slide data tailored to a specific instruction is a non-trivial task. For instance, the instruction ``Fix all typos in the slide'' requires the slide to actually contain typographical errors, while ``Translate the slide into Chinese'' assumes the slide is written in another language which is not a Chinese.

To generate appropriate slides for each of the 379 instructions, we observed that each seed instruction and its GPT-4o-generated variants are associated with the same base PowerPoint file \citep{openai2024gpt4o}.
Based on this insight, we manually created slide decks for each of the 56 seed instructions. To ensure visual quality and realism, we adopted publicly available templates% from Microsoft
\footnote{https://create.microsoft.com/en-us/search?filters=powerpoint}
and we listed 10 template which we used in benchmark dataset in Table~\ref{tab:template_links}. 

We release the full benchmark, including metadata that maps each instruction to its corresponding PowerPoint file and instruction group. A summary of this mapping is presented in Table~\ref{tab:tsbench_categories}, and detailed statistics are provided in Appendix~\ref{benchmark-statistics}.

\paragraph{TSBench-Hard}
While the core TSBench focuses on explicit and modular editing tasks, real-world user interaction is often underspecified or requires contextual reasoning. Since PowerPoint is an inherently visual medium, purely language-driven agents may struggle with tasks requiring spatial awareness or aesthetic inference. To systematically evaluate these operational boundaries, we introduce a challenging subset named \textit{TSBench-Hard}.
This subset comprises instructions in three high-difficulty categories:
(1)~\textit{Visual-Dependent Tasks} that require spatial reasoning impossible to resolve with text matching alone (e.g., ``Align the text box to the left edge of the image'');
(2)~\textit{Ambiguous Instructions} where the agent must infer the user's aesthetic intent from vague directives (e.g., ``Make the title slide look more professional''); and
(3)~\textit{Impossible or Cross-Modal Tasks} that require perception inaccessible to a non-visual agent, evaluating its ability to correctly refuse execution rather than hallucinate (e.g., ``Identify the speaker in the embedded video'').
By incorporating these edge cases, TSBench-Hard serves as a stress test to determine the operational boundaries of slide editing agents.
Illustrative examples are provided in Appendix~\ref{sec:appendix_hard_examples}.
% \begin{figure}
%     \centering
%     \includegraphics[width=1\linewidth]{cost_fig.pdf}
%     \caption{Visualization of the average cost per instruction across different instruction categories for each system.}

%     \label{fig:cost}
% \end{figure}

\section{Experiment}
In this section, we evaluate our proposed system, \textsc{Talk-to-Your-Slides}, against two baseline methods using the benchmark dataset we introduced. Our evaluation measures both traditional performance and efficiency metrics, such as total cost. Each baseline is designed to address a specific research question.

\textbf{For RQ1}, we evaluate a GUI-based agent baseline by including \textsc{UFO2}\citep{zhang2024ufo}. This agent is capable of operating a wide range of Windows applications, including PowerPoint, and utilizes the Gemini-2.5-flash model. Additional configuration details are provided in Appendix~\ref{app:ufo}.

\input{table-model-comparation}

\textbf{For RQ2}, we establish a direct code generation baseline. This method leverages our document understanding module (Section~\ref{Parser}) to extract a structured representation of each slide. It then prompts an LLM with this representation and the user instruction to generate executable code.

\paragraph{Configurations}Our proposed framework utilizes two core large language models: the instruction understanding module is powered by Gemini-1.5-flash, while the document editing and code generation modules use Gemini-2.5-flash. The code generator is configured with up to three retry attempts. For comparison, we also conducted experiments by replacing this default Gemini setup with other models, specifically GPT-4.1-mini~\citep{openai2023gpt4}, Claude-haiku~\citep{anthropic2024claude3}, and DeepSeek V3~\citep{deepseekai2024deepseekv3technicalreport}. While the main results presented in this paper are based on our primary Gemini configuration, a detailed analysis of the results from these other LLMs is available in Appendix~\ref{app:auxiliary-results}. Full model descriptions and hyperparameter settings are provided in Appendix~\ref{app:models}.

\paragraph{Performance metrics}
To evaluate the performance of the slide editing LLM agent, we introduce three main metrics. First, we adopt the \textit{Execution Success Rate} (SR) from prior work~\citep{ zhang2024pptcrbenchmarkevaluatingrobustness}. SR indicates whether the finally generated code is successfully executed. Second, we use \textit{LLM judge scores}, also employed in~\citep{zheng2025pptagentgeneratingevaluatingpresentations, wang2025can}. These scores encompass text, image, layout, and color aspects. To ensure the reliability of the LLM judge on these four aspects, we also conducted a human evaluation (Appendix~\ref{app:human-eval}). In addition to these, we added \textit{instruction following} metric, which indicates how well the edited presentation reflects the user's command. Our third metric is \textit{Execution Time}, which refers to the latency in seconds required to execute a single instruction. More specific configurations are provided in Appendix~\ref{app:details-of-evaluation}.

\textbf{Efficiency metrics}
To assess system efficiency, particularly in the context of \textbf{\textit{RQ1}}, we include three metrics: \textit{Average Input Tokens}, \textit{Average Output Tokens}, and \textit{Average Cost}. All averages are calculated for a single instruction. Token counts are computed by aggregating all tokens passed to the LLM within each module (including all four modules), and the cost is then estimated accordingly. Detailed information, including the API pricing for each LLM, is described in Appendix~\ref{app:details-of-evaluation}.

\section{Results}
\label{Results}
In this section, we will answer the research questions mentioned in the Section~\ref{introduction} based on the experimental results above.
Then based on these findings, we discuss how agents should approach tasks that assist humans beyond slide editing when interacting with computing systems in general.
\subsection{Main Results}
This section answers two main research questions. First (RQ1), we investigate whether it is necessary to use an image modality from screen captures, as in conventional methods, to control a program. Second (RQ2), we explore if a level-wise architecture allows an LLM agent to operate an application more effectively. In response, we present the following research answers:

\paragraph{(RA1) Language can be used to control applications as an alternative to visual input.}
This means that an LLM agent does not require GUI screen captures as input; instead, it can operate using the internal language that functions within the program. As shown in Table~\ref{tab:main_table}, our system demonstrates superior performance on average when compared to the UI Agent system, both in Deepseek and Gemini. Although the UI Agent was more proficient on LayoutAdjustment instructions, which require significant visual confirmation, our system reduced the execution time to 66\% of the UI Agent's.
Moreover from an efficiency perspective, while our system generated more output tokens, it achieved an overwhelming advantage by using only 4\% of the input tokens. Consequently, the average cost was lowered to just 13\% of the UI Agent's.

\paragraph{(RA2) A level-wise architecture LLM agent can successfully operate an application.}
Our experiments show that an LLM agent that divides its workflow into high-level and low-level processes achieves a distinct performance improvement over direct code generation as shown in Table~\ref{tab:main_table}. The level-wise architecture, designed based on the nature of user-computer interaction explained in Section~\ref{method}, scaled at test time. This led to a 160\% increase in the success rate and a marked improvement in other performance metrics. This signifies that we successfully achieved test-time scaling~\citep{muennighoff2025s1} by reconfiguring the LLM agent's architecture.

To justify the reliability of the LLM judge scores, we conducted an experiment on 30 instances, which yielded a Pearson correlation coefficient above 0.8 and similarly high Spearman correlation across all judge metrics as shown in Table~\ref{tab:correlations-gemini}.
Four illustrative examples of slides edited by \textsc{Talk-to-Your-Slides} are shown in Figure~\ref{example_fig}, and additional results conducted with other LLMs are in Appendix~\ref{app:auxiliary-results}.

\subsection{Comparing the LLMs}
We find that the choice of the underlying language model involves distinct trade-offs. For example, comparing Table~\ref{tab:main_table} DeepSeek V3 (0324) and Gemini-2.5-flash, we see that in the Direct code generation baseline, Gemini-2.5-flash achieved a success rate of 59.90\%, which was lower than DeepSeek's 77.30\%. However, when leveraged within our system, Gemini-2.5-flash's performance improved to 96.83\%, effectively matching GPT-4.1-mini's 96.84\%. Although the detailed evaluations for Text, Image, Layout, and Color were similar between the two, our system achieved state-of-the-art results more frequently when using DeepSeek V3 (0324). From an efficiency perspective, although DeepSeek generated fewer tokens (1.87k) compared to Gemini-2.5-flash (2.53k), its higher unit pricing meant that it was relatively more expensive per token, despite the lower total cost (\$1.4 vs \$2.0).

Furthermore, as summarized in Table~\ref{tab:model_comparison} (which shows results for our system exclusively), Gemini-2.5-flash used more input and output tokens than the other models. The higher input token count is likely due to token amplification as data passes through the four modules of our system. Notably, this resulted in a significantly higher instruction following score compared to the other models.
Additionally, to demonstrate the practical robustness of our method in large-scale scenarios, we provide a full batch inference example processing an 89-page slide deck in Appendix~\ref{batch_slide_example}.

Meanwhile, could we create an even more efficient and capable LLM agent by supplementing the language modality with a vision modality? We discuss this topic in Appendix~\ref{argument1}.

\subsection{TSBench-Hard Results}
\label{sec:tsbench-hard-results}
\input{appendix_hard_result}

Table~\ref{tab:tsbench-hard-main} reports execution success rates (SR) and refusal accuracy (RA) on TSBench-Hard. Given the high complexity of these tasks, existing baselines struggle significantly; the \textit{Direct code generation} baseline achieves a near-zero success rate on visually dependent tasks, highlighting its inability to ground code in the visual state of slides.

Our system achieves the highest performance across most categories, with an overall SR of $31.3\%$, outperforming the strongest baseline (UI Agent) by $+16.5$ points. Notably, our system demonstrates superior capability in detecting infeasible requests, achieving a Refusal Accuracy of $64.7\%$, suggesting stronger semantic understanding of task boundaries.

However, the results also reveal the \textbf{operational boundaries} of our non-visual approach: on Visual-Dependent tasks, the UI Agent achieves a comparable SR ($12.8\%$ vs.\ our $12.5\%$), confirming that tasks requiring spatial reasoning remain challenging without visual perception. In contrast, our system substantially outperforms on Ambiguous ($29.5\%$ vs.\ $18.2\%$) and Multi-step ($24.1\%$ vs.\ $10.4\%$) tasks, where structured data access and hierarchical reasoning provide clear advantages.
An ablation on self-reflection and detailed retry statistics are provided in Appendix~\ref{app:self-relfection-example}.

\subsection{State Representation vs.\ Code Generation}
\label{sec:state-rep}
A key distinction of our approach is that it operates on the document's \textit{state representation}---the underlying object model (DOM)---rather than generating code from scratch to render visual output. Unlike code-driven image generation methods (e.g., TikZ-based approaches~\citep{belouadi2023automatikz}), our system \textit{edits} existing objects by modifying their properties (position, text, color) through API calls, ensuring that all elements remain fully editable native PowerPoint objects.

This design provides an advantage for batch processing: by operating on structured data, unmentioned elements are guaranteed to remain untouched, achieving the structural fidelity required for professional workflows. Re-generating slides via code rendering often leads to unintended layout shifts or hallucinated content. A more detailed comparison is provided in Appendix~\ref{sec:appendix_tikz_comparison}.

\subsection{Discussion: Toward Hybrid Agents}
\label{sec:discussion-hybrid}
Our results paint a clear picture of complementary strengths. Language-driven structural agents excel at text-centric, batch, and multi-step editing with high efficiency and low cost. GUI-based agents, while slower and more expensive, remain competitive on spatially dependent tasks. We therefore position our system not as a total replacement for GUI agents, but as a \textit{high-efficiency module for structured tasks} that could sit within a hybrid architecture. In such a system, structured manipulation handles the majority of editing operations, while visual perception is selectively invoked for layout verification and aesthetic quality assurance.

\section{Conclusion}
We introduced \textsc{Talk-to-Your-Slides}, an LLM-powered agent for editing slides.
By decomposing editing tasks into level-wise object manipulations, our system interacts with application in language modality to execute complex edits.
We developed \textit{TSBench}, a benchmark with 379 diverse editing instructions, including a \textit{Hard} subset that systematically probes the operational boundaries of both visual and non-visual agents.
Experiments show our approach significantly outperforms baselines on text-centric and batch editing tasks, demonstrating the effectiveness of a language-driven structural approach.

\section*{Limitations}

By prioritizing code-based manipulation over visual processing, our agent achieves significantly higher speed, lower cost, and precise structural fidelity. However, this design choice introduces several concrete trade-offs.

\paragraph{Layout stability.} When text length changes substantially---for instance, translating Chinese to English often doubles character count---the resulting text may overflow its bounding box. Detecting such overflow from structured data alone requires computing precise glyph extents and comparing them against container geometry, which is non-trivial without visual feedback (see the concrete example in Appendix~\ref{app:future_work}).

\paragraph{Visual aesthetics.} Subjective instructions such as ``Make this slide look balanced'' or ``Shrink the image to a proper size'' require spatial perception that is inaccessible to a non-visual agent. As shown in TSBench-Hard (Table~\ref{tab:tsbench-hard-main}), our system's success rate on Visual-Dependent tasks ($12.5\%$) is comparable to the UI Agent ($12.8\%$), confirming that both paradigms struggle---but for different reasons.

These limitations reinforce our positioning of the system as a \textit{high-efficiency module for structured tasks} rather than a complete replacement for GUI agents. We believe the most promising direction is a \textit{Hybrid Agent} architecture (Section~\ref{sec:discussion-hybrid}), where our efficient structural manipulation handles the majority of batch tasks, while VLM capabilities are selectively invoked for layout verification and visual quality assurance.

\section*{Ethics Statement}
\label{ethical-considerations}
Our work introduces \textsc{Talk-to-your-Slides}, a system designed to enhance productivity and accessibility in slide editing, and TSBench, a benchmark for evaluating such systems. We hope our contributions will foster further research into language-driven software agents that reduce tedious manual labor.

We acknowledge potential sources of bias in our work. The \textit{TSBench} dataset, while curated for quality, is built upon 56 seed instructions that were augmented using GPT-4o. Although the topics are diverse, covering areas from marketing to linguistics, the initial instructions and content may reflect the perspectives of their creators and may not represent all possible user domains. To mitigate this, all 379 instructions in the final benchmark underwent a manual human review process to filter for clarity, validity, and unambiguous objectives.

Furthermore, our agent relies on large language models (LLMs) such as Gemini and GPT, which may inherit societal biases from their training data. This could potentially influence how the agent interprets instructions or modifies content. In addition, our human evaluation was conducted with 21 participants noted as Appendix~\ref{app:human-eval}. While this provided valuable qualitative feedback, the limited sample size may not fully capture the diverse range of user perspectives, and the findings should be interpreted with this in mind. We recommend that users, especially when working with sensitive topics, review the agent's output.

Since the system processes user slide data, which could be confidential, users should be mindful of the data privacy policies of the underlying LLM APIs. We encourage the responsible use of this technology as a tool to assist, not replace, human oversight.

\section*{Acknowledgements}
This work was supported by Institute for Information \& communications Technology Planning \& Evaluation(IITP) grant funded by the Korea government(MSIT) (RS-2019-II190075, Artificial Intelligence Graduate School Program(KAIST)).

This work was supported by the National Research Foundation of Korea(NRF) grant funded by the Korea government(MSIT) (No. RS-2025-00555621)

This work was supported by the Institute of Information \& Communications Technology Planning \& Evaluation(IITP) grant funded by the Korea government(MSIT) (RS-2025-02304967, AI Star Fellowship(KAIST))

\bibliography{main}

\clearpage
%\onecolumn

\appendix

% --- Appendix Overview Box ---
% --- Updated Appendix Overview Box ---
\section*{Appendix}
We provide detailed supplementary materials organized as follows:

\begin{itemize}[leftmargin=1.5em, label={}]
    \item \textbf{Appendix~\ref{app:doc-understanding}} and \textbf{Appendix~\ref{app:self-relfection-example}} detail the system implementation, including document parsing and self-reflection mechanisms.
    \item \textbf{Appendix~\ref{sec:appendix_tikz_comparison}} compares our approach with code-driven image generation methods.
    \item \textbf{Appendix~\ref{app:understanding output}} provides example of output in understanding modules.
    \item \textbf{Appendix~\ref{app:detail-of-tsbench}} provides comprehensive statistics and details of the TSBench dataset, including Hard subset examples and evaluation details.
    \item \textbf{Appendix~\ref{app:models}} and \textbf{Appendix~\ref{app:ufo}} describe the experimental models and the UFO baseline.
    \item \textbf{Appendix~\ref{app:alternative}} details our primary implementation on Windows via COM and discusses AppleScript as a macOS alternative.
    \item \textbf{Appendix~\ref{app:details-of-evaluation}} and \textbf{Appendix~\ref{app:auxiliary-results}} present evaluation details (including human eval) and auxiliary experimental results.
    \item \textbf{Appendix~\ref{argument1}} and \textbf{Appendix~\ref{batch_slide_example}} discuss the role of GUI images and present qualitative results on batch slide inference.
    \item \textbf{Appendix~\ref{app:future_work}} discusses future directions, specifically addressing the challenges of fine-grained layout adjustment and the necessity of visual feedback.
    \item \textbf{Appendix~\ref{app:prompt}} lists the full prompts used for planning, editing, coding, and evaluation.
\end{itemize}
\hrulefill
\vspace{1em}

% -----------------------------
\section{Details of document understanding}
\label{app:doc-understanding}

Although a \texttt{.pptx} file is technically a collection of XML files, directly working with these raw XML structures is highly impractical for document-level understanding. The XML files are excessively verbose, and their content is fragmented across multiple interlinked components. This makes parsing both time-consuming and error-prone.

XML format of PowerPoint is also inherently index-based. For example, attributes such as text formatting (e.g., bold, italic) are not stored as human-readable strings but instead encoded as numeric indices that point to style definitions in separate lookup tables. As a result, even simple queries such as ``is this text bolded?'' require multi-step resolution across files.

To mitigate these challenges, \citet{zheng2025pptagentgeneratingevaluatingpresentations} proposed converting slides into HTML format. While this method simplifies parsing to some extent, it introduces an additional rendering step and often fails to retain the full range of visual and structural information available in the original slide-particularly layout metadata, positional precision, and fine-grained formatting.

In contrast, our system avoids HTML-based conversion and instead directly parses the PowerPoint object model through COM (Component Object Model) interfaces. This design choice enables precise extraction of layout, style, and object-level data with full fidelity to the original file. The extracted information is then normalized into a structured JSON format to support downstream semantic reasoning and editing tasks.

\section{Detailed Analysis of Self-Reflection Mechanism}
\label{app:self-relfection-example}
In this section, we provide a concrete example to elaborate on the self-reflection mechanism described in Section 3.4. The self-reflection module is triggered when the execution layer encounters a runtime error (e.g., \texttt{com\_error} in Windows or API failures). Upon detecting an error, the agent analyzes the full error trace, reflects on the syntax or logical flaws, and regenerates the code. This iterative process continues until the code executes successfully or reaches a pre-defined maximum number of attempts (\texttt{max\_attempt\_num}).

We present a real-world scenario in Figure~\ref{fig:appendix_reflection_case} where the agent resolves a data type mismatch during a PowerPoint automation task.

\subsection{Impact of Self-Reflection}
We further investigate the contribution of the self-reflection mechanism in Table~\ref{tab:tsbench-hard-retry}. The ablation study reveals that self-reflection is critical for handling complex instructions.
\paragraph{Performance Gain:} Enabling self-reflection yields a consistent improvement across all sub-categories, boosting the overall SR from $23.5\%$ to $31.3\%$ ($+7.8\%$). The gain is particularly pronounced in Multi-step tasks ($+8.8\%$), where iterative refinement helps correct intermediate logic errors.
\paragraph{Error Mitigation:} More importantly, self-reflection drastically reduces the Catastrophic Failure (CF) rate from $24.8\%$ to $8.7\%$. Without reflection, the agent often generates code that corrupts slide elements or throws unhandled exceptions. The reflection loop acts as a safety guard, catching these errors before they finalize.

\begin{figure}[h!]
    \centering
    \begin{minipage}{0.48\textwidth}
    \fbox{
        \begin{minipage}{0.95\textwidth}
            \fontsize{10}{10}\selectfont
            %%%%%%%%%
            % Title
            \textbf{Case Study: Self-Correction Loop} \\
            \rule{\textwidth}{0.4pt}
            \fontsize{9}{10}\selectfont  
            %%%%%%%%%%
            % Body content start
            \textbf{User Instruction:} "Change the title color to red."
            
            \vspace{0.2cm}
            \textbf{1. Attempt 1 (Generated Code):} \\
            \texttt{\# Agent incorrectly tries to assign a tuple} \\
            \texttt{slide.Shapes(1)...Font.Color.RGB = (255, 0, 0)}
            
            \vspace{0.1cm}
            % Error in Red
            \textcolor{red}{\textbf{Execution Error:}} \\
            \textit{TypeError: Objects of type 'tuple' can not be converted to a COM VARIANT}
            
            \vspace{0.2cm}
            \textbf{2. Refinement Step (Self-Reflection):} \\
            "I attempted to assign a tuple (255, 0, 0) directly. However, the PowerPoint COM interface expects a single integer... I must convert it."
            
            \vspace{0.2cm}
            \textbf{3. Attempt 2 (Regenerated Code):} \\
            \texttt{\# Agent corrects the format to an integer} \\
            \texttt{slide.Shapes(1)...Font.Color.RGB = 255}
            
            %%%%%%%%%%%
            % Body content end
            \vspace{0.1cm}
            \rule{\textwidth}{0.4pt}
            \fontsize{9}{10}\selectfont
            %%%%%%%%
            % Footer / Result in Blue
            \textbf{Result}: \textcolor{blue}{\textbf{Success}} (Color updated to red)
        \end{minipage}
        }
    \end{minipage}
    \caption{A real-world example of the self-reflection mechanism handling a data type error during execution.}
    \label{fig:appendix_reflection_case}
\end{figure}

\input{selfreflection_ablation}

\section{Comparison with Code-Driven Image Generation}
\label{sec:appendix_tikz_comparison}

The Area Chair and reviewers noted similarities between our code-generation approach and existing text-to-image generation works, such as TikZ-based methods (e.g., Automatikz\citep{belouadi2023automatikz}). While both paradigms utilize LLMs to generate code for visual outputs, there are fundamental differences in their objectives, operational mechanics, and practical applications.

\paragraph{Object Manipulation vs. Pixel Rendering.}
The primary goal of TikZ-based approaches is \textit{generation}—rendering a static image or vector graphic from scratch based on a description~\citep{belouadi2023automatikz, ellis2017tikz}. In contrast, our work targets \textit{editing} and \textit{manipulation} of existing documents. We do not render pixels; instead, we modify the properties (e.g., position, text, color) of pre-existing objects within the slide's Document Object Model (DOM). This distinction is critical: generating a slide as a single image (or TikZ compilation) results in a non-editable, \textit{dead} output where individual elements cannot be easily modified by the user afterwards. Our API-based approach ensures that all elements remain fully ``alive'' and editable native PowerPoint objects.

\paragraph{Structural Fidelity in Batch Processing.}
As emphasized in our positioning, our key contribution lies in \textit{efficient text-centric batch processing}. Image generation models (visual or code-based) often struggle to maintain strict structural fidelity across multiple slides (e.g., ``Change the font of all titles to Arial while keeping their exact positions''). Re-generating the entire slide via TikZ or VLM often leads to unintended layout shifts or hallucinations. By strictly operating on the underlying structured data (XML/COM), our method guarantees that unmentioned elements remain untouched, achieving the high fidelity and low latency required for professional, large-scale workflows that image generation methods cannot support.

\paragraph{Operational Efficiency.}
Rendering images or compiling complex TikZ code is computationally expensive and slow. Our approach, which executes lightweight API calls to modify attributes, bypasses the rendering pipeline entirely. This efficiency is what enables the massive cost and time reductions (as shown in Table~\ref{tab:main_table}) compared to visual approaches, making it the only viable solution for real-time, bulk editing tasks.

\section{Example of Understanding Output}
\label{app:understanding output}
\begin{figure}[t]
    \centering
    \includegraphics[width=0.8\columnwidth, trim={290 190 290 190}, clip]{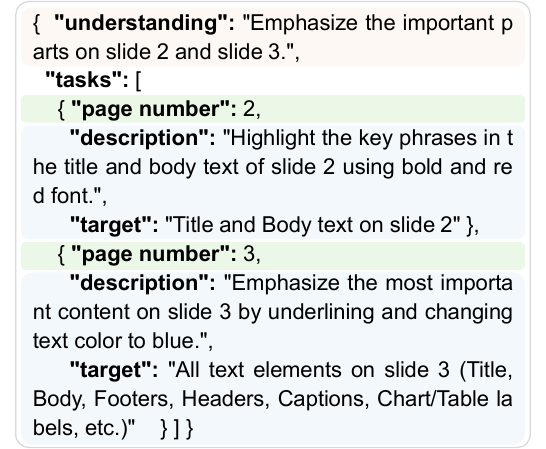}
    \caption{Example output generated by the instruction understanding module.}
    \label{fig:planner}
\end{figure}

\begin{figure}[t]
    \centering
    \includegraphics[width=0.8\linewidth, trim={270 70 270 75}, clip]{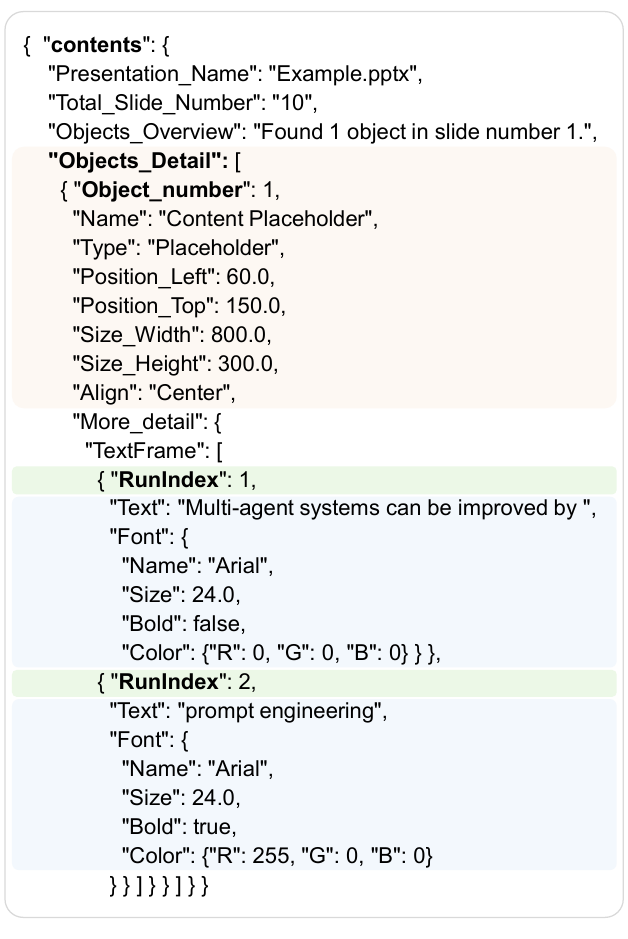}
    \caption{Example output of document understanding. The yellow sections contain information about the parsed object's name, type, location, size, and other details. The runs highlighted in green demonstrate that different text formatting styles can exist within a single text box.}
    \label{fig:parser}
\end{figure}

In this section, we show the example output generated by instruction understanding module (Figure~\ref{fig:planner} and document understanding module (Figure~\ref{fig:parser}.

\section{Detail of TSBench}
\label{app:detail-of-tsbench}

In this section, we present detailed information about TSBench. First, we have listed the topics of slide content in Table~\ref{tab:content topics}.

Additionally, the mapping between instruction numbers, slide IDs, and the four categories is listed in Table~\ref{tab:tsbench_categories}. Each instruction is assigned an \texttt{instruction\_key} of the form \texttt{n} or \texttt{n-m}, where \texttt{n} denotes one of the original 56 seeds, and \texttt{n-m} indicates the 
$m$th augmentation derived from seed \texttt{n}. The PowerPoint files are named \texttt{slide\_<instruction\_key>.pptx} (e.g., \texttt{slide\_0.pptx}, \texttt{slide\_3-1.pptx}), with the suffix matching the corresponding \texttt{instruction\_key}.

The list of slide templates is enumerated in Table~\ref{tab:template_links} along with their corresponding links.
\input{appendix_slide_topic}
\input{appendix_instruction-category-map}

\begin{figure*}[t]
    \centering
    \includegraphics[width=1\linewidth]{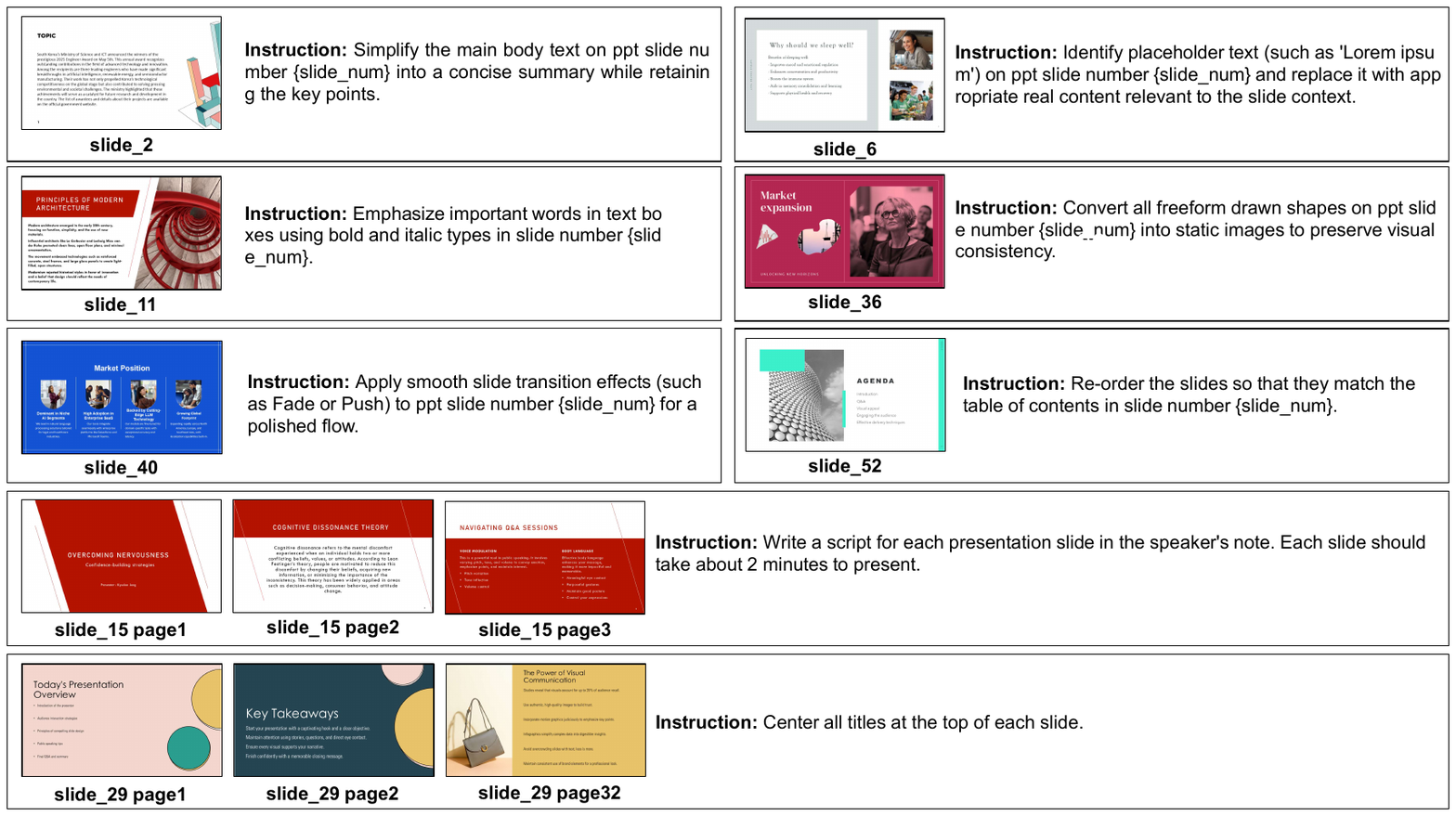}
    \caption{Example from the TSBench dataset. Some data points consist of a single slide, while others contain multiple slides.}
    \label{fig:TSBench-example}
\end{figure*}

Examples of TSBench slides and instructions are presented in Figure~\ref{fig:TSBench-example}.

\input{appendix_tsbench_hard_example}
\subsection{Statistics}
\label{benchmark-statistics}
Table~\ref{tab:instruction_stats} reports the number of instructions in each of the four categories. In the \textit{Total} column, we observe that the original set of 56 human‐authored seed instructions was expanded to 560 through GPT-4o‐based augmentation. After excluding 181 examples with unclear objectives or those deemed unsuitable for benchmarking, a final set of 379 instruction–slide pairs remained. Consequently, TSBench comprises 379 instructions and 56 corresponding \texttt{.pptx} files. Detailed information, including the mapping between instructions and \texttt{.pptx} files, is provided in Table~\ref{tab:tsbench_categories}, and slide content topics are listed in Table~\ref{tab:content topics}.
\input{table_benchmark_stat}

\subsection{Detail of TSBench-Hard}
\label{sec:appendix_hard_examples}
The three categories of TSBench-Hard (Visual-Dependent, Ambiguous, and Impossible/Cross-Modal) are described in the main text (Section~4). Here we provide illustrative examples in Figure~\ref{fig:tsbench_hard_examples}.

\begin{figure}[t]
    \centering
    \includegraphics[width=0.8\linewidth]{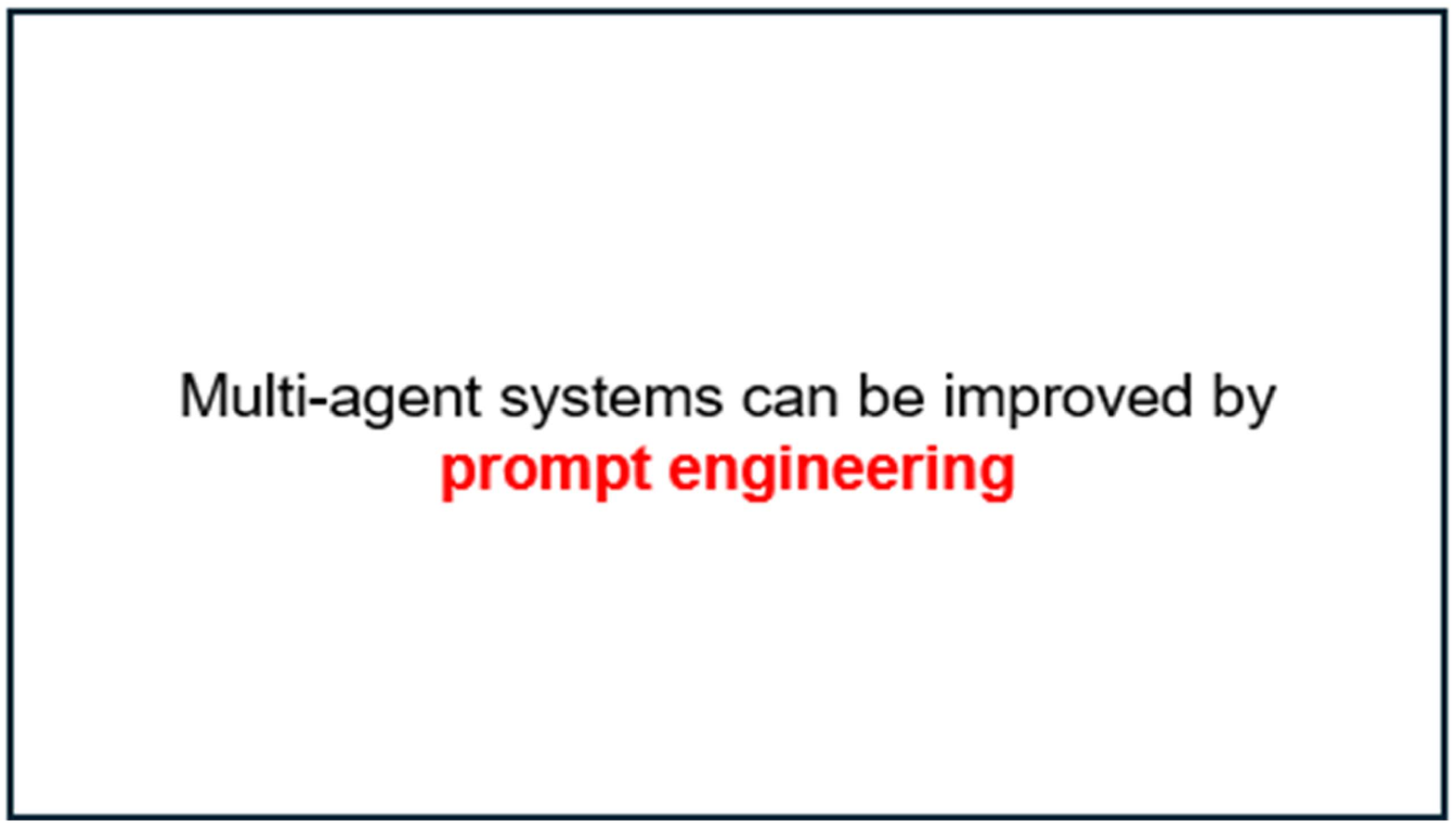}
    \caption{The original slide from which the example parsed data was extracted.}
    \label{fig:parser-image}
\end{figure}

In addition, the original slide is presented in Figure~\ref{fig:parser-image} from which the example parsed data Figure~\ref{fig:parser} was extracted.

\subsection{TSBench-Hard Evaluation Details}
The main TSBench-Hard results are reported in Section~\ref{sec:tsbench-hard-results}.
For this evaluation, the image of the edited slide was fed into \texttt{gemini-3-flash}. The model rated the extent to which the result satisfied the dataset's \texttt{ideal\_description} on a scale of 0 to 5, which was subsequently converted into a percentage.

\section{Model}
\label{app:models}

In this section, we provide details about the external LLM APIs used in the experiments. Gemini-2.5-flash and GPT-4.1-mini were embedded in the agent system, while GPT-4o was used as a judge when evaluating the system. The price of each models is illustrated in Table~\ref{tab:model-detail}.

\paragraph{Gemini-1.5-flash (gemini-1.5-flash)}
Gemini 1.5 Flash~\citep{comanici2025gemini25pushingfrontier} offers cost-effective multimodal capabilities with tiered pricing based on context length. Released in early 2024, this model represents Google's focus on balancing accuracy with efficiency. It supports a context window of up to 1,048,576 tokens, max output token is 8,192, allowing for processing of extensive documents and conversations in a single request. The model handles multiple modalities including text, code, images, and limited audio processing. While not as powerful as its larger counterparts in complex reasoning tasks, it demonstrates strong capability in straightforward instruction following, summarization, and multimodal understanding tasks. We used this model for instruction understanding with maxtoken: 2048, temperature 0.05, and top\_p 1.0.

\input{appendix_ppt_template}

\paragraph{Gemini-2.5-flash (gemini-2.5-flash-preview-04-17).}
Gemini 2.5 Flash~\citep{comanici2025gemini25pushingfrontier} is a high-throughput thinking model designed to strike an optimal balance between speed, cost, and reasoning capability. As Google's latest preview-tier model, it extends the popular 2.0 Flash foundation with major upgrades in reasoning capability, while still prioritizing low latency and economical usage for developers. It supports a wide range of modalities—including text, code, images, audio, and video—making it well suited for diverse AI tasks where both multimodal understanding and cost efficiency are critical. We used this model for document editing and code generation with the maximum supported token limit of 65536, a temperature of 0.05, and top\_p of 1.0.

\paragraph{GPT-4.1-mini (gpt-4.1-mini-2025-04-14).}  
GPT-4.1-mini~\citep{openai2023gpt4} is the `mini' variant of the GPT-4.1 family, released April 14, 2025. It inherits the core strengths of the flagship GPT-4.1 series—state-of-the-art coding ability, robust instruction following, and support for very long (up to one million-token) contexts—while reducing model size to cut inference latency by roughly 50\% and lower operational cost. This makes GPT-4.1-mini an ideal choice for applications that demand the latest model capabilities in a more resource-efficient footprint. We used this model for document editing and code generation with the maximum supported token limit of 32768, a temperature of 0.05, top\_p 1.0.

\paragraph{GPT-4o (gpt-4o-2024-08-06).}  
GPT-4o (``o'' for ``omni'')~\citep{openai2023gpt4} is OpenAI's multimodal flagship, released August 6, 2024. It can ingest and generate text, images, and audio in real time, enabling unified reasoning across these modalities. Compared to its predecessor (GPT-4 Turbo), GPT-4o offers faster API throughput and lower per-token cost, making it especially powerful for tasks that require seamless cross-modal understanding and generation. We used this model as an LLM judge with a max token of 512, a temperature of 0.2, top\_p 1.0.

\paragraph{Claude 3 Haiku (claude-3-haiku-20240307).}
Claude 3 Haiku~\citep{anthropic2024claude3} is the fastest and most economical model in Anthropic's Claude 3 family. It is designed for near-instantaneous responsiveness, making it ideal for applications requiring real-time interaction, such as customer support chatbots and content moderation. Like the other models in the Claude 3 series (Opus and Sonnet), Haiku possesses robust vision capabilities and supports a 200K token context window. The model is positioned as a cost-effective solution for enterprise workloads where speed and accuracy are critical, offering strong performance relative to other models in its class.
\paragraph{DeepSeek-V3-0324}
DeepSeek-V3~\citep{deepseekai2024deepseekv3technicalreport} is a powerful Mixture-of-Experts (MoE) language model comprising 671 billion total parameters, with 37 billion activated for each token. It leverages the Multi-head Latent Attention (MLA) and DeepSeekMoE architectures, previously validated in DeepSeek-V2, to ensure efficient inference and cost-effective training. Innovatively, DeepSeek-V3 introduces an auxiliary-loss-free strategy for load balancing and a multi-token prediction training objective to enhance performance. The model was pre-trained on a high-quality dataset of 14.8 trillion tokens, followed by Supervised Fine-Tuning (SFT) and Reinforcement Learning (RL). Evaluations show that DeepSeek-V3 surpasses other open-source models and rivals the performance of leading closed-source models, while requiring a notably efficient 2.788 million H800 GPU hours for its complete and stable training process.
\input{appendix_model_price}

\section{UFO: GUI Agent}
\label{app:ufo}

We employ \textsc{UFO2}~\citep{zhang2024ufo}, a state-of-the-art multi-agent GUI automation system for Windows desktops, as one of our baseline methods. UFO2 is designed to execute natural language instructions by integrating both visual and API-based control over various applications. It features a centralized \textit{HostAgent} for task decomposition and orchestration, and multiple \textit{AppAgents} tailored to specific applications such as PowerPoint.

In our experimental setup, both the HostAgent and AppAgent are powered by the vision-language model Gemini-2.5-flash. The system is further backed by a retry mechanism with up to three fallback attempts to ensure robust execution. This configuration enables UFO2 to leverage its hybrid GUI–API action interface, speculative multi-action execution, and a Picture-in-Picture (PiP) sandbox for non-intrusive user experience.

For additional technical details and evaluation results, we refer readers to the original UFO2 paper~\citep{zhang2024ufo}.

\begin{figure}[t]
    \centering
    \includegraphics[width=1\columnwidth, trim={260 180 260 190}, clip]{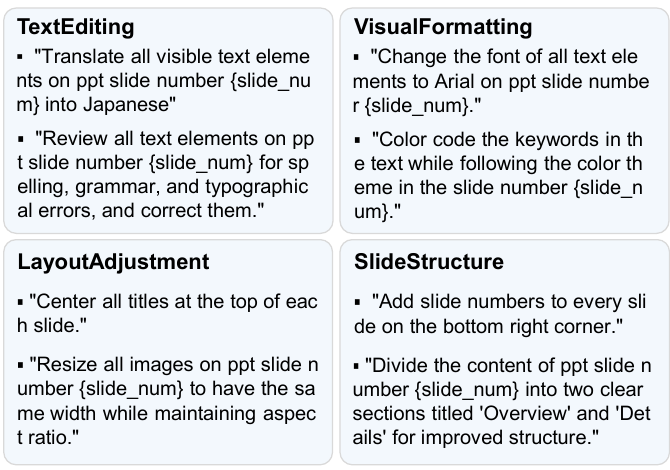}
    \caption{Examples of instructions across four categories.}
    \label{inst_example}
\end{figure}

\input{appendix_com_applescript}

\section{Details of evaluation}
\label{app:details-of-evaluation}
Instruction following measures how effectively the 
The remaining four metrics are established as reference-free metrics following \citep{ge2025autopresentdesigningstructuredvisuals}. Scores range from 0 (worst) to 5 (best), with detailed criteria ensuring consistent interpretation.
We employ the multimodal \texttt{gpt-4o} model for this evaluation.
The model assesses edit quality by comparing original and edited slide images along with slide notes and instructions. The full scoring prompt appears in Figure~\ref{app-fig:judge-prompt} and ~\ref{app-fig:judge-TILC-prompt}.

\subsection{Human evaluation}
\label{app:human-eval}
We conducted a human evaluation with 21 volunteers (17 male, 4 female) from Chung-Ang University, all with extensive experience using PowerPoint slides in both academic and external settings. Participants evaluated a subset of 30 items from our benchmark. Similar to the LLM judge, the volunteers were asked to rate the results on a scale from 0 (worst) to 5 (best) based on four criteria: text, image, layout, and color. The correlation coefficient presented in our analysis is calculated based on these 30 data points. None of the participants were color-blind. The ethical considerations for this study are detailed in Section~\ref{ethical-considerations}.

Table~\ref{tab:main_table} reports these metrics using Gemini-2.5-flash in its non-thinking output mode, with cost computed based on the pricing as of May 16, 2025.\footnote{As of 2025-05-16: \$0.15 per million input tokens (text, image, video), \$1.00 per million input tokens (audio), \$0.60 per million output tokens (non-thinking), \$3.50 per million output tokens (thinking).}

\section{Auxiliary results}
\label{app:auxiliary-results}
In this section, we report the results of our experiments conducted using gpt-4.1-mini~\citep{openai2023gpt4}, Claude-haiku~\citep{anthropic2024claude3}. The results are presented in Table~\ref{tab:gpt4_mini_results} and Table~\ref{tab:claude_haiku_results}, the correlation coefficients for the metrics assessed by the LLM judge are reported in Table~\ref{tab:correlations-gpt}.

We also report the human correlation results for the Gemini-2.5-flash experiments, where model outputs were evaluated using LLM-based judges. These correlation analyses are presented in Figure~\ref{tab:correlations-gemini}.

\input{appdendix_result_gpt-4.1-mini}
\input{judge-human-corr-gpt4-1-mini}

\input{appendix_result_claude_haiku}

\section{Should Software Agents Ultimately Use Only GUI Images?}
\label{argument1}
To better understand the trade-offs between GUI-based and code-based approaches in slide editing, we present concrete examples using the same instructions applied through different methods.
When comparing the efficiency of GUI-based and code-based slide editing approaches, errors in VLM-based optical character recognition (OCR) provide persuasive evidence. The GUI agent, which relies on Vision Language Models (VLMs) for text recognition, failed to accurately identify text due to OCR limitations, resulting in low scores on text editing metrics. In contrast, the code-based system \textsc{Talk-to-Your-Slides} directly accessed text data without requiring VLM-based OCR processing. These results demonstrate that direct access to structured data is more reliable than external perception mechanisms like VLMs, highlighting how system accuracy depends on data accessibility.

Nevertheless, visual information from the GUI remains useful in software automation. There are editing tasks where purely low-level textual information is insufficient. For instance, when translating Chinese text into English, the translated content often expands in length, causing overflow beyond the original text box. In such cases, visual layout information helps preserve the slide’s aesthetic quality. Here, GUI images complement the limitations of low-level approaches and can raise the upper bound of what can be achieved.

In this work, we demonstrated that a low-level, code-based approach is both effective and efficient for many core editing tasks in presentation software. Moving forward, we believe future research should explore hybrid approaches that combine the semantic precision and efficiency of structured parsing with the contextual awareness of visual understanding. We encourage the community to pursue this direction to further advance automated presentation editing systems.

\section{Examples of Batch Slide Inference Results}
\label{batch_slide_example}
This section presents examples of actual lecture materials (originally in Korean) translated into English using Talk-to-your-slides. We have processed large-scale PowerPoint decks, specifically \textit{07\_RNN\_LSTM\_Seq2Seq} (89 pages) and \textit{08\_Transformer} (49 pages), and made the results publicly available at the following link: \url{https://drive.google.com/drive/folders/1-TmujhQkMg17QsdFx_LOCVRr19QNFyOd?usp=sharing}.

\begin{figure*} \centering \includegraphics[width=1\linewidth]{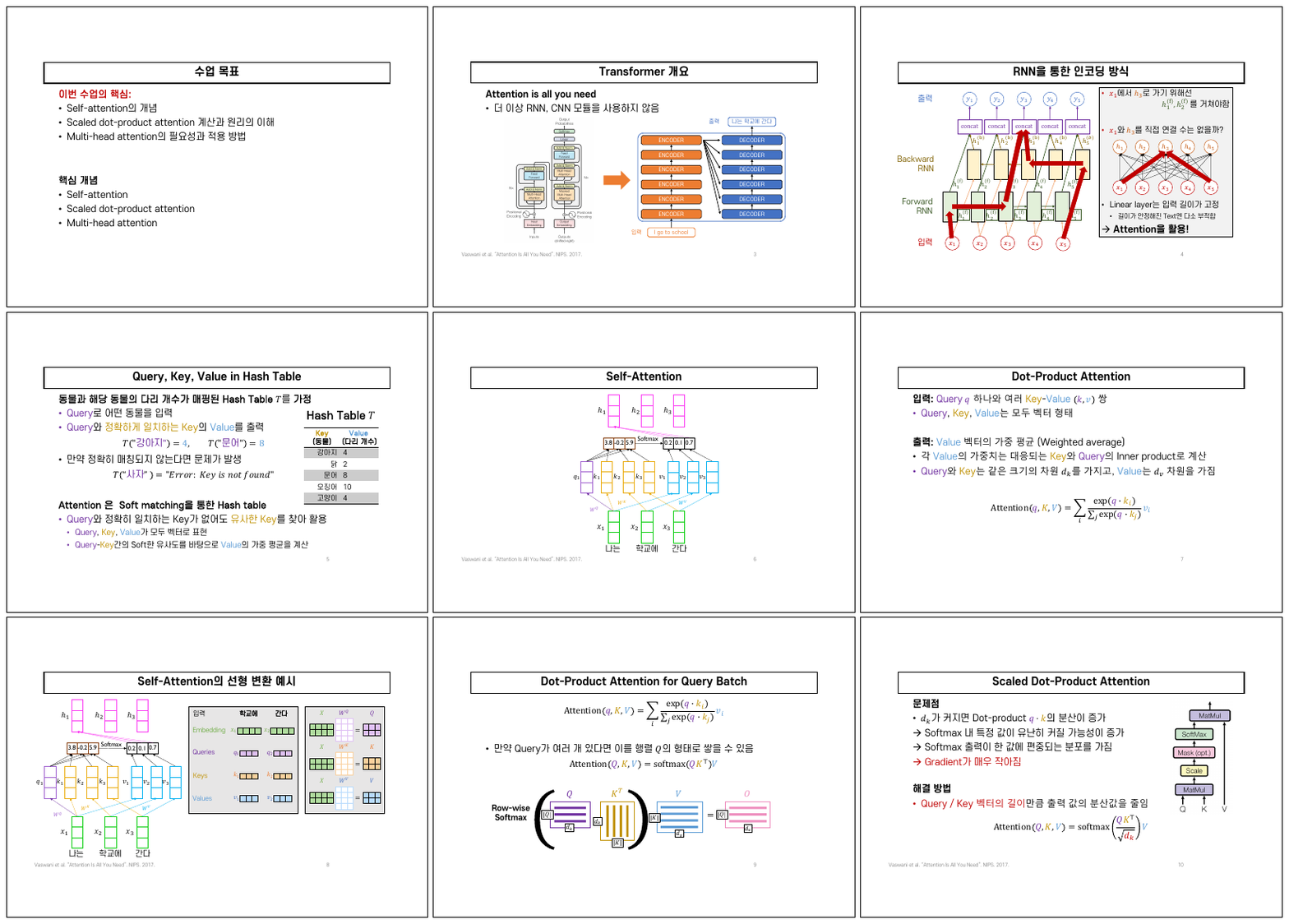} \caption{Example of the original lecture slide before processing. This is a part of the batch editing process (editing 89 slides). You can find the file in \href{https://docs.google.com/presentation/d/1QrfHnwRCYDtDdyk-eCT3oXiPDEeIKTFD/edit?usp=share_link&ouid=113126270543442786981&rtpof=true&sd=true}{\colorbox{winnerblue}{here}}.} \label{fig:batch_example_before} \end{figure*}

\begin{figure*} \centering \includegraphics[width=1\linewidth]{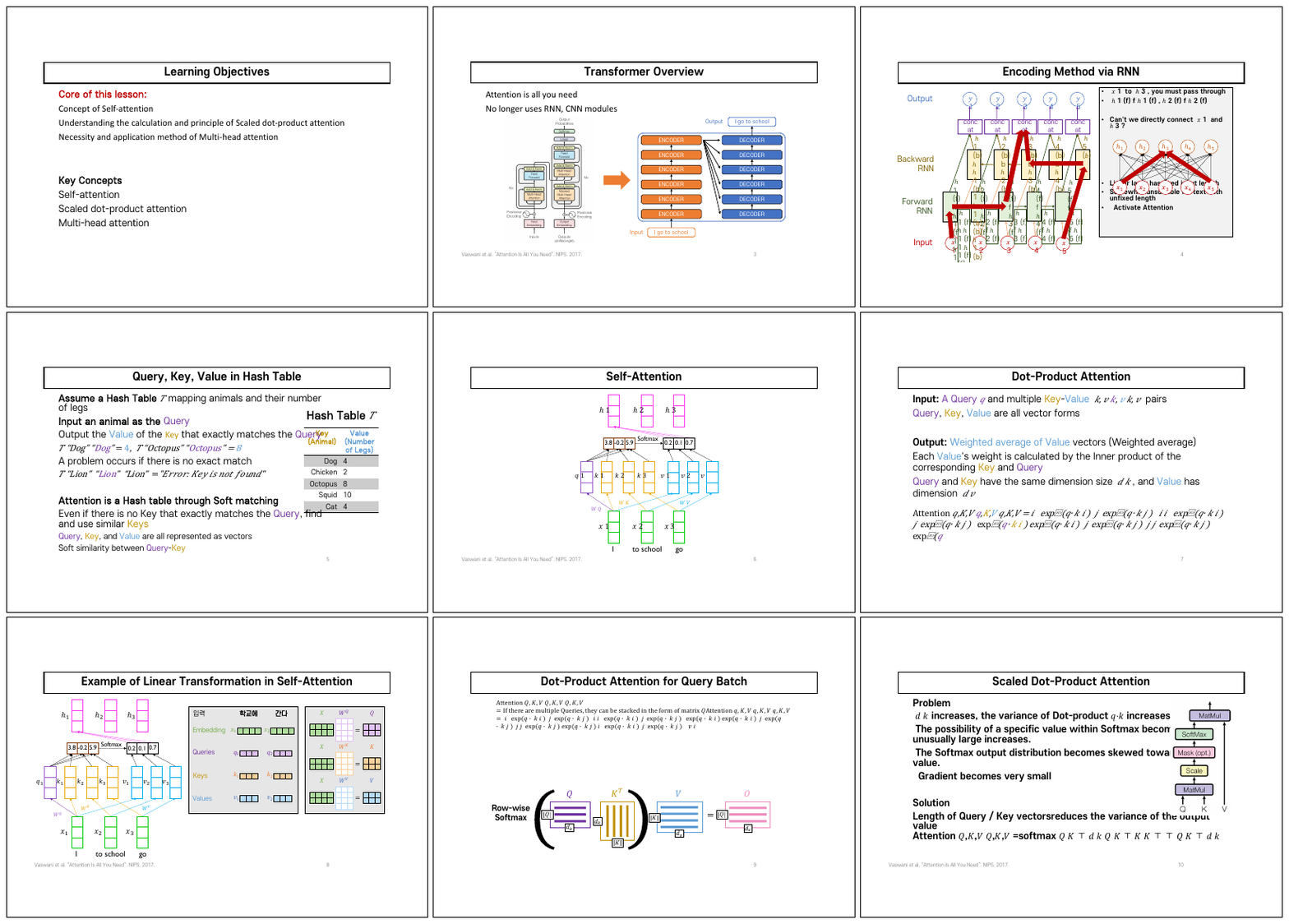} \caption{Example of the original lecture slide after processing. This is a part of the batch editing process (editing 89 slides). You can find the file in\href{https://docs.google.com/presentation/d/1Kl6Kkru7UtHifK4QwJ473Ox11WjuM63_/edit?usp=share_link&ouid=113126270543442786981&rtpof=true&sd=true}{\colorbox{winnerblue}{here}}.} \label{fig:batch_example_after} \end{figure*}

Figure~\ref{fig:batch_example_before} illustrates the original PPT slides, while Figure~\ref{fig:batch_example_after} shows the processed versions. The translation into English was executed successfully. Notably, as shown in the slide at row 2, column 1, terms with the same meaning were successfully edited at the `run' level while maintaining consistent coloring.

However, some limitations remain for future work. First, mathematical formulas are occasionally altered by the LLM into illegible formats. Second, in cases involving complex illustrations or diagrams, the transition from Korean to English can lead to text box overflows due to differences in text length and structure.

\section{Future Work}
\label{app:future_work}
Research in slide editing, such as the methods discussed in Section~\ref{related:slide-generation} and our proposed \textsc{Talk-to-your-Slides}, is still in its early stages. This raises a critical question from a practical user perspective: \textit{\textbf{what is the most significant area for improvement?}} We propose that the primary challenge lies in handling fine-grained layout adjustments.

\begin{figure}[h]
    \centering
    \fbox{\includegraphics[width=1.0\linewidth]{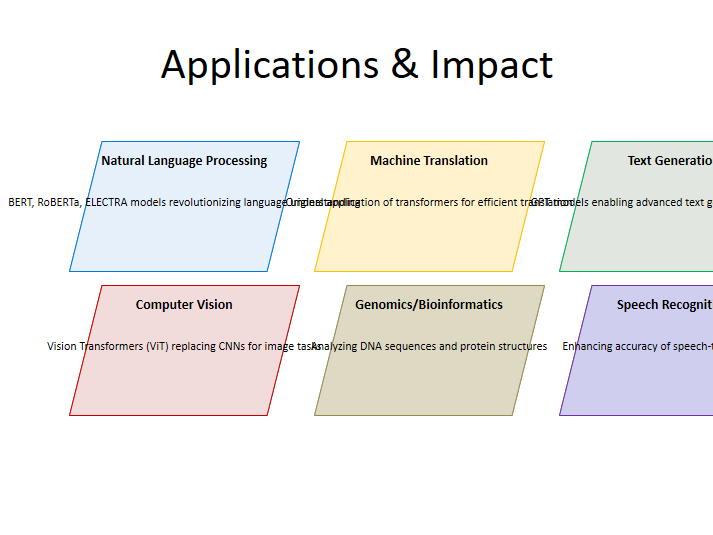}}
    \caption{An example of a layout error in the slide editing results. (see Appendix~\ref{app:MCP})}
    \label{fig:error-case}
\end{figure}

As illustrated in Figure~\ref{fig:error-case}, while the title, shape, sub-caption, and text are correctly inserted, the layout of the text box is improper. A human user editing this slide would naturally resize the text box to fit the text neatly within the parallelogram. However, determining this text overflow using only PowerPoint's XML data is a non-trivial task. It would require calculating the precise boundaries of the parallelogram and comparing them with the coordinates of the text box to judge whether an overflow has occurred. This process is computationally complex.

For such spatial reasoning tasks, it would be far more effective for the agent to perceive the slide visually, for instance, through a screenshot. As we will further discuss in Appendix~\ref{argument1}, we believe the ideal slide editing agent, the ultimate goal for our research community, should be a language-based model that leverages vision as an auxiliary input to determine its actions.

\section{Prompt}
\label{app:prompt}
In this section, we present the prompts used for the LLMs in our experiments. Following \citet{Chen_2025} and \citet{liu-etal-2024-lost}, we carefully designed detailed prompts. As experimental outcomes can vary significantly depending on subtle differences in prompts, we disclose our prompts in full to ensure specificity and reproducibility.

\subsection{instruction understanding prompt}
\label{planner_prompt}
In the instruction understanding stage, the system interprets the user's intent and formulates a plan \citep{aghzal2025surveylargelanguagemodels, oelerich2024language, hao2024planning}. The prompt of this is shown in Figure~\ref{app-fig:planner-prompt}.
\input{appendix_planner_prompt}
%\newpage
\subsection{document editing prompt}
\label{processor_prompt}
In the document editing stage, the system generates the post-editing data in JSON format based on the plan and the parsed data. This process is illustrated in Figure~\ref{app-fig:document-editing-prompt}.
\input{appendix_processor_prompt}
%\newpage
\subsection{Code generator prompt}
In the code generation stage, the system takes as input the original slide data, the document-edited slide data, and the plan, and outputs Python code that applies the corresponding changes in PowerPoint. The prompt used for this step is shown in Figure~\ref{app-fig:code-generator-prompt}.
\label{applier_prompt}

\input{appendix_applier_prompt}

%\newpage
\subsection{Direct code generation prompt}
\label{baseline_prompt}
In the case of the Direct code generation, the system directly generates code using only the parsed slide data and the instruction, without intermediate planning or editing. The prompt used for this process is shown in Figure~\ref{app-fig:baseline-prompt}.
\input{appendix_baseline_prompt}
%\newpage

\subsection{LLM judge prompt}
\label{app:judge_prompt}
To evaluate slide editing capabilities, we employed an LLM-based judge, following a similar approach to \citet{ge2025autopresentdesigningstructuredvisuals}. The prompt used to assess how well the instruction was followed is shown in Figure~\ref{app:judge_prompt}, while the prompt used for evaluating text, image, layout, and color,based on the criteria from \citet{ge2025autopresentdesigningstructuredvisuals}, is presented in Figure~\ref{app-fig:judge-TILC-prompt}.
\input{appendix_judge_prompt}
\label{TILC_prompt}
\input{appendix_judge_TILC_prompt}

\end{document}

%% file: main_table_edit.tex
\definecolor{winnerblue}{RGB}{220, 240, 255} 
\definecolor{badred}{RGB}{255, 235, 235}     

\begin{table*}[t]
\centering
\tiny
\renewcommand{\arraystretch}{0.85}
\setlength{\tabcolsep}{3.5pt}
\resizebox{\textwidth}{!}{%
\begin{tabular}{@{} l l c c c c c c c r r r @{} }
\toprule
\textbf{System} & \textbf{Instruction} & \multicolumn{7}{c}{\textbf{Performance metric} ($\uparrow$)} & \multicolumn{3}{c}{\textbf{Efficiency metric} ($\downarrow$)} \\
\cmidrule(lr){3-9} \cmidrule(lr){10-12}
 & & \textbf{SR (\%)} & \makecell[c]{\textbf{Inst.}\\\textbf{Foll.}} & \textbf{Text} & \textbf{Image} & \textbf{Layout} & \textbf{Color} & \makecell[c]{\textbf{Exec.}\\\textbf{Time (s)}} & \makecell[c]{\textbf{Avg.}\\\textbf{In (k)}} & \makecell[c]{\textbf{Avg.}\\\textbf{Out (k)}} & \makecell[c]{\textbf{Cost}\\\textbf{($\times$10$^{-3}$)}} \\

% ==========================================================================================
% SECTION 1: DeepSeek V3 (0324)
% ==========================================================================================
\midrule[\heavyrulewidth]
\multicolumn{12}{c}{\textbf{\textit{Model A: DeepSeek V3 (0324)}}} \\
\midrule

\multirow{5}{*}{\shortstack[l]{Direct \\ code gen.}}
 & TextEditing & 77.05 & 0.00 & 0.00 & 1.01 & 1.02 & 1.02 & \underline{16.30} & \underline{1.00} & \underline{0.39} & \underline{0.4} \\
 & VisFmt      & 79.02 & 0.54 & 0.75 & 1.04 & 1.03 & 0.87 & \underline{15.92} & 1.12 & \underline{0.43} & \underline{0.5} \\
 & LayoutAdj   & 70.10 & 0.08 & 1.47 & 1.08 & 1.41 & 1.41 & \underline{15.20} & 1.02 & \underline{0.51} & \underline{0.5} \\
 & SlideStruct & 85.02 & 0.18 & 1.80 & 1.12 & 1.93 & 1.66 & \underline{17.10} & \underline{0.89} & \underline{0.62} & \underline{0.6} \\
 & \textbf{Overall} & 77.30 & 0.55 & 0.83 & 1.06 & 1.25 & 1.16 & \underline{16.13} & \underline{1.03} & \underline{0.46} & \underline{0.5} \\
\midrule
\multirow{5}{*}{UI Agent}
 & TextEditing & 66.10 & 0.49 & 0.69 & 1.60 & 1.50 & 1.57 & 101.30 & 97.80 & 2.05 & 5.3 \\
 & VisFmt      & 88.75 & 2.46 & 1.91 & 1.68 & 1.81 & 1.50 & 122.40 & 90.10 & 2.38 & 5.8 \\
 & LayoutAdj   & 63.02 & 1.49 & 2.78 & \textbf{2.21} & 2.25 & 2.47 & 114.10 & 116.00 & 2.40 & 6.7 \\
 & SlideStruct & 82.74 & \textbf{1.56} & 2.28 & 1.32 & 2.10 & 2.01 & 88.30 & 73.10 & 1.72 & 4.2 \\
 & \textbf{Overall} & \cellcolor{badred}75.65 & \cellcolor{badred}1.70 & \cellcolor{badred}1.97 & \cellcolor{badred}1.55 & \cellcolor{badred}2.18 & \cellcolor{badred}2.04 & \cellcolor{badred}114.03 & \cellcolor{badred}94.80 & \cellcolor{badred}2.14 & \cellcolor{badred}5.5 \\
\midrule
\multirow{5}{*}{\textbf{Ours}}
 & TextEditing & \textbf{98.55} & \textbf{2.62} & \textbf{3.03} & \textbf{3.11} & \textbf{3.57} & \textbf{3.62} & \textbf{52.70} & \textbf{3.30} & \textbf{1.63} & \textbf{1.2} \\
 & VisFmt      & \textbf{96.10} & 2.02 & \textbf{2.05} & \textbf{1.73} & \textbf{2.27} & \textbf{1.90} & \textbf{82.30} & \textbf{4.18} & \textbf{2.11} & \textbf{1.6} \\
 & LayoutAdj   & \textbf{98.02} & 1.41 & 2.39 & 1.83 & \textbf{2.28} & \textbf{2.49} & \textbf{82.10} & \textbf{3.92} & \textbf{2.06} & \textbf{1.5} \\
 & SlideStruct & \textbf{91.70} & 1.41 & \textbf{2.46} & \textbf{2.61} & \textbf{2.81} & \textbf{2.92} & \textbf{69.00} & \textbf{2.20} & \textbf{1.26} & \textbf{0.8} \\
 & \textbf{Overall} & \cellcolor{winnerblue}\textbf{96.84} & \cellcolor{winnerblue}\textbf{2.12} & \cellcolor{winnerblue}\textbf{2.48} & \cellcolor{winnerblue}\textbf{2.27} & \cellcolor{winnerblue}\textbf{2.73} & \cellcolor{winnerblue}\textbf{2.69} & \cellcolor{winnerblue}\textbf{71.55} & \cellcolor{winnerblue}\textbf{3.55} & \cellcolor{winnerblue}\textbf{1.87} & \cellcolor{winnerblue}\textbf{1.4} \\

% ==========================================================================================
% SECTION 2: Gemini-2.5-flash
% ==========================================================================================
\midrule[\heavyrulewidth]
\multicolumn{12}{c}{\textbf{\textit{Model B: Gemini-2.5-flash}}} \\
\midrule

\multirow{5}{*}{\shortstack[l]{Direct \\ code gen.}}
 & TextEditing & 62.07 & 0.00 & 0.10 & 1.70 & 1.70 & 1.70 & \underline{23.08} & \underline{1.21} & \underline{0.93} & \underline{0.7} \\
 & VisFmt      & 53.66 & 0.44 & 0.84 & 1.04 & 1.04 & 0.89 & \underline{37.72} & \underline{1.38} & 2.75 & 1.9 \\
 & LayoutAdj   & 58.95 & 0.66 & 1.55 & 0.79 & 1.42 & 1.40 & \underline{23.25} & \underline{1.26} & \underline{0.94} & \underline{0.8} \\
 & SlideStruct & 73.33 & 0.50 & 1.61 & 1.25 & 1.58 & 1.72 & \underline{28.17} & \underline{1.00} & \underline{1.05} & \underline{0.8} \\
 & \textbf{Overall} & 59.90 & 0.36 & 0.88 & 1.22 & 1.41 & 1.37 & \underline{28.35} & \underline{1.24} & \underline{1.48} & \underline{1.0} \\
\midrule
\multirow{5}{*}{UI Agent}
 & TextEditing & 66.38 & 0.54 & 0.50 & 1.63 & 1.49 & 1.79 & 117.85 & 102.04 & 2.26 & 16.6 \\
 & VisFmt      & 86.18 & \textbf{2.61} & 2.01 & 1.72 & 2.25 & 1.80 & 122.25 & 93.27 & 2.15 & 15.2 \\
 & LayoutAdj   & 65.26 & \textbf{2.07} & \textbf{2.82} & \textbf{2.38} & \textbf{2.51} & \textbf{2.87} & 128.23 & 108.48 & 2.50 & 17.7 \\
 & SlideStruct & 82.22 & 1.67 & \textbf{2.31} & 1.55 & \textbf{2.38} & 2.17 & 99.18 & 72.46 & 1.82 & 12.0 \\
 & \textbf{Overall} & \cellcolor{badred}74.41 & \cellcolor{badred}1.64 & \cellcolor{badred}1.81 & \cellcolor{badred}1.83 & \cellcolor{badred}2.11 & \cellcolor{badred}2.10 & \cellcolor{badred}119.66 & \cellcolor{badred}97.29 & \cellcolor{badred}2.23 & \cellcolor{badred}15.9 \\
\midrule
\multirow{5}{*}{\textbf{Ours}}
 & TextEditing & \textbf{99.14} & \textbf{2.95} & \textbf{3.01} & \textbf{2.65} & \textbf{3.11} & \textbf{3.07} & \textbf{55.98} & \textbf{3.81} & \textbf{1.96} & \textbf{1.6} \\
 & VisFmt      & \textbf{94.30} & 1.98 & \textbf{2.22} & \textbf{1.86} & \textbf{2.38} & \textbf{2.16} & \textbf{94.78} & \textbf{5.09} & \textbf{3.58} & \textbf{2.8} \\
 & LayoutAdj   & \textbf{100.0} & 1.80 & 2.39 & 2.15 & 2.37 & 2.56 & \textbf{86.14} & \textbf{4.46} & \textbf{2.29} & \textbf{2.0} \\
 & SlideStruct & \textbf{91.10} & \textbf{1.71} & 1.95 & \textbf{1.73} & 2.17 & \textbf{2.35} & \textbf{96.54} & \textbf{2.71} & \textbf{1.69} & \textbf{1.2} \\
 & \textbf{Overall} & \cellcolor{winnerblue}\textbf{96.83} & \cellcolor{winnerblue}\textbf{2.21} & \cellcolor{winnerblue}\textbf{2.48} & \cellcolor{winnerblue}\textbf{2.17} & \cellcolor{winnerblue}\textbf{2.58} & \cellcolor{winnerblue}\textbf{2.57} & \cellcolor{winnerblue}\textbf{78.95} & \cellcolor{winnerblue}\textbf{4.26} & \cellcolor{winnerblue}\textbf{2.53} & \cellcolor{winnerblue}\textbf{2.0} \\
\bottomrule
\end{tabular}%
}
\caption{Cross-model comparison results. `SR' denotes execution success rate. \textbf{Ours} (highlighted in \colorbox{winnerblue}{blue}) achieves the best balance of high performance and efficiency. While `Direct Code Gen' shows lower latency (underlined), its significantly lower SR makes it impractical. Cost is in USD ($\times$10$^{-3}$).}
\label{tab:main_table}
\end{table*}

%% file: human_eval_performance_gemini.tex
\definecolor{lightblue}{RGB}{233,243,250}
\definecolor{pastelred}{RGB}{252,228,236}
\begin{table}[t]
%\begin{minipage}{\textwidth}
\centering
\resizebox{0.5\textwidth}{!}{%
\begin{tabular}{@{} l l c c c c @{} }
\toprule
\textbf{System}
& \textbf{Instruction}
& \multicolumn{4}{c}{\textbf{Human Evaluation}} \\
\cmidrule(l){3-6}
&
& \makecell[c]{\textbf{Text}}
& \makecell[c]{\textbf{Image}}
& \makecell[c]{\textbf{Layout}}
& \makecell[c]{\textbf{Color}} \\
\midrule
\multirow{6}{*}{\shortstack[l]{Direct \\ code\\ generation}}
& TextEditing & 1.55 & 2.20 & 2.25 & 2.10 \\
& VisualFormatting & 1.88 & 1.95 & 1.90 & 1.82 \\
& LayoutAdjustment & 2.11 & 1.76 & 2.05 & 2.01 \\
& SlideStructure & 2.05 & 1.91 & 2.15 & 2.24 \\
& Overall & 1.90 & 1.96 & 2.09 & 2.04 \\
\cmidrule(l){2-6}
& \textbf{PCC} & 0.84 & 0.69 & 0.73 & 0.88 \\
\midrule
\multirow{6}{*}{UI Agent}
& TextEditing & 2.45 & 2.88 & 2.79 & 2.91 \\
& VisualFormatting & 3.01 & 2.72 & 3.25 & 2.80 \\
& LayoutAdjustment & 3.85 & 3.41 & 3.55 & 3.76 \\
& SlideStructure & 3.31 & 2.55 & 3.38 & 3.17 \\
& Overall & 3.16 & 2.89 & 3.24 & 3.16 \\
\cmidrule(l){2-6}
& \textbf{PCC} & 0.87 & 0.69 & 0.71 & 0.8 \\
\midrule
\multirow{6}{*}{Ours}
& TextEditing & 4.15& 3.95 & 4.22 & 4.18 \\
& VisualFormatting & 3.22 & 2.86 & 3.38 & 3.16 \\
& LayoutAdjustment & 3.39 & 3.15 & 3.37 & 3.56 \\
& SlideStructure & 3.95 & 2.73 & 3.17 & 3.35 \\
& Overall & 3.68 & 3.17 & 3.54 & 3.56 \\
\cmidrule(l){2-6}
& \textbf{PCC} & 0.89 & 0.75 & 0.72 & 0.94 \\
\bottomrule
\end{tabular}%
}
%\end{minipage}
\caption{Human evaluation of slide editing performance using Gemini-2.5-flash with Pearson Correlation Coefficient (PCC) against an LLM judge. All $p$-values are below $10^{-3}$.}
\label{tab:correlations-gemini}
\end{table}

%% file: table-model-comparation.tex
\begin{table}[t]
\centering
\resizebox{0.5\textwidth}{!}{%
\begin{tabular}{lccccc}
\toprule
\textbf{Model} & \textbf{SR (\%)} & \makecell[c]{\textbf{Instr.}\\ \textbf{Follow}} & \makecell[c]{\textbf{Overall}\\ \textbf{Perf.}} & \makecell[c]{\textbf{Exec.}\\ \textbf{Time (s)}} & \makecell[c]{\textbf{Avg.}\\ \textbf{Cost (\$)}} \\
\midrule
Gemini-2.5-flash & 96.83 & \textbf{2.21} & \textbf{2.48} & 78.95 & 0.0020 \\
GPT-4.1-mini & 96.57 & 2.13 & 2.46 & 78.37 & 0.0038 \\
Claude 3.5 Haiku & 95.89 & 2.08 & 2.41 & 75.33 & 0.0033 \\
DeepSeek V3 (0324) & \textbf{96.84} & 2.12 & \textbf{2.48} & \textbf{71.55} & \textbf{0.0014} \\
\bottomrule
\end{tabular}%
}
\caption{Cross-model comparison of Ours across Gemini-2.5-flash, GPT-4.1-mini, Claude 3.5 Haiku, and DeepSeek V3. Efficiency metrics are averaged across tasks. Cost in USD.}
\label{tab:model_comparison} % Added a label for cross-referencing
\end{table}

%% file: appendix_hard_result.tex
% =========================================================
% TSBench-Hard: Result templates (drop-in LaTeX)
% Requires: \usepackage{booktabs, multirow, makecell, xcolor, siunitx}
% Optional: \usepackage{pifont} for checkmarks, or remove.
% =========================================================

% ---------- (A) Color defs (same style as your tables) ----------
\definecolor{lightblue}{RGB}{220, 240, 255}

% ---------- (B) Suggested paragraph templates (copy & fill numbers) ----------
% Put this in the TSBench-Hard subsection right before/after the table.

% \paragraph{TSBench-Hard setup.}
% We introduce \textsc{TSBench-Hard}, a curated subset designed to stress-test (i) visually dependent edits,
% (ii) ambiguous aesthetic requests, (iii) long-horizon multi-step logic, and (iv) impossible requests that
% require safe refusal. We report execution-based success (SR, \%), and for the ``Impossible'' category we
% additionally report refusal accuracy (RA, \%), measuring whether the agent correctly identifies and declines
% infeasible actions.

% \paragraph{Main results.}
% Table~\ref{tab:tsbench-hard-main} shows that \textbf{Ours} consistently improves SR on text-centric and
% multi-step edits, while remaining competitive on visually dependent tasks where purely language-based access
% is fundamentally limited. Notably, on ``Impossible'' requests, \textbf{Ours} achieves the highest RA,
% indicating safer failure behavior.

% \paragraph{Self-reflection ablation.}
% Enabling self-reflection improves SR by \textbf{+X.X} points overall, with the largest gains on multi-step
% and visually dependent instructions, where runtime API errors and format mismatches are common.
% We also observe a reduction in catastrophic failures (CF) from \textbf{Y.Y\%} to \textbf{Z.Z\%}.

% ---------- (C) Table 1: TSBench-Hard category breakdown (SR / RA) ----------
\begin{table}[h]
\centering
\resizebox{0.49\textwidth}{!}{%
\begin{tabular}{@{} l c c c c c @{}}
\toprule
\textbf{System} &
\makecell[c]{\textbf{Visual-}\\\textbf{Dependent}\\\textbf{SR}\,(\%)} &
\makecell[c]{\textbf{Ambiguous}\\\textbf{SR}\,(\%)} &
\makecell[c]{\textbf{Multi-step}\\\textbf{SR}\,(\%)} &
\makecell[c]{\textbf{Impossible}\\\textbf{RA}\,(\%)} &
\makecell[c]{\textbf{Overall}\\\textbf{SR}\,(\%)} \\
\midrule
Direct code generation & 4.1 & 11.6 & 8.9 & 22.4 & 9.2 \\
UI Agent & \textbf{12.8} & 18.2 & 10.4 & 35.0 & 14.8 \\
\rowcolor{lightblue}
\textbf{Ours} & 12.5 & \textbf{29.5} & \textbf{24.1} & \textbf{64.7} & \textbf{31.3} \\
\bottomrule
\end{tabular}%
}
\caption{\textbf{TSBench-Hard results.} SR: execution success rate. RA: refusal accuracy on infeasible requests
(higher is better). Best results are highlighted in bold.}
\label{tab:tsbench-hard-main}
\end{table}

%% file: selfreflection_ablation.tex
% ---------- (D) Table 2: Self-reflection ablation on TSBench-Hard ----------
\begin{table}[h]
\centering
\resizebox{0.49\textwidth}{!}{%
\begin{tabular}{@{} l c c c c c @{}}
\toprule
\textbf{Variant} &
\makecell[c]{\textbf{Visual-}\\\textbf{Dependent}\\\textbf{SR}\,(\%)} &
\makecell[c]{\textbf{Ambiguous}\\\textbf{SR}\,(\%)} &
\makecell[c]{\textbf{Multi-step}\\\textbf{SR}\,(\%)} &
\makecell[c]{\textbf{Overall}\\\textbf{SR}\,(\%)} &
\makecell[c]{\textbf{CF}\,(\%) } \\
\midrule
Ours (w/o self-reflection) & 20.4 & 21.1 & 15.3 & 23.5 & 24.8 \\
\rowcolor{lightblue}
\textbf{Ours (w/ self-reflection)} & \textbf{26.8} & \textbf{29.5} & \textbf{24.1} & \textbf{31.3} & \textbf{8.7} \\
\bottomrule
\end{tabular}%
}
\caption{\textbf{Ablation on self-reflection.} CF: catastrophic failure rate (e.g., unhandled exception,
corrupted slide state, or repeated invalid actions). Define CF consistently with your pipeline logs.}
\label{tab:tsbench-hard-ablation}
\end{table}

% ---------- (E) Table 3: Retry statistics (optional but very persuasive) ----------
\begin{table}[h]
\centering
\resizebox{0.49\textwidth}{!}{%
\begin{tabular}{@{} l c c c c @{}}
\toprule
\textbf{System} &
\makecell[c]{\textbf{Avg.}\\\textbf{\#Attempts}} &
\makecell[c]{\textbf{P90}\\\textbf{\#Attempts}} &
\makecell[c]{\textbf{Refine}\\\textbf{Triggered}\,(\%)} &
\makecell[c]{\textbf{Latency}\\\textbf{Overhead}\,(\%)} \\
\midrule
Ours (w/o self-reflection) & 1.00 & 1 & 0.0 & -- \\
\rowcolor{lightblue}
\textbf{Ours (w/ self-reflection)} & 1.42 & 3 & 45.6 & +38.2 \\
\bottomrule
\end{tabular}%
}
\caption{\textbf{Self-reflection efficiency.} Attempt counts and overhead on TSBench-Hard.
``Refine Triggered'' is the fraction of instances that entered the self-correction loop at least once.}
\label{tab:tsbench-hard-retry}
\end{table}

% ---------- (F) Minimal text snippet to reference these tables ----------
% You can paste this verbatim and just fill in numbers.
%
\noindent\textbf{TSBench-Hard.} Table~\ref{tab:tsbench-hard-main} reports performance on visually dependent,
ambiguous, multi-step, and impossible instructions. Our agent improves overall SR by \textbf{+16.5} points over
the strongest baseline, while achieving the best refusal accuracy (RA) on infeasible requests. As shown in
Table~\ref{tab:tsbench-hard-ablation}, self-reflection yields consistent gains (overall \textbf{+7.8} SR) and
reduces catastrophic failures, at a modest runtime overhead (Table~\ref{tab:tsbench-hard-retry}).

%% file: appendix_slide_topic.tex
\begin{table*}[h!]
\centering
\caption{Content topics of slides in TSBench dataset.}
\label{tab:tsbench_topics}
\resizebox{0.8\textwidth}{!}{%
\begin{tabular}{@{}cllcllcll@{}}
\toprule
\textbf{Index} & \textbf{Topic} && \textbf{Index} & \textbf{Topic} && \textbf{Index} & \textbf{Topic} \\
\midrule
0 & Linguistics && 19 & Economics && 38 & AI Strategy \\
1 & Communication && 20 & Climate Change && 39 & Marketing \\
2 & News Articles && 21 & Quotes && 40 & Marketing \\
3 & Presentation && 22 & Some Numbers && 41 & Marketing \\
4 & Remote Work && 23 & Aesthetics && 42 & Company \\
5 & Data Collection && 24 & Presentation && 43 & Company \\
6 & Sleeping && 25 & Presentation && 44 & Marketing \\
7 & LLMs && 26 & Design && 45 & Financials \\
8 & Hypertension && 27 & Q\&A && 46 & Competitive Landscape \\
9 & Impressionism && 28 & Presentation && 47 & Product Overview \\
10 & Immanuel Kant && 29 & Visual Communication && 48 & AI Assistant Platform \\
11 & Modern Architecture && 30 & Presentation Theme && 49 & Future Outlook \\
12 & Aesthetics && 31 & What I Like && 50 & Design \\
13 & Education && 32 & What I Like && 51 & Presentation \\
14 & Cognitive Dissonance && 33 & Creative Vision && 52 & Visual Appeal \\
15 & Nervousness && 34 & Sports && 53 & Design \\
16 & Linear Algebra && 35 & Marketing Strategies && 54 & NLP \\
17 & Artificial Intelligence && 36 & Marketing && 55 & Presentation \\
18 & Economics && 37 & Marketing && & \\
\bottomrule
\label{tab:content topics}
\end{tabular}
}
\end{table*}

%% file: appendix_instruction-category-map.tex
\begin{table*}[t]
\centering
\caption{Category mapping of slides in TSBench dataset.}
\label{tab:tsbench_categories}
\resizebox{0.8\textwidth}{!}{%
\begin{tabular}{@{}cllcllcll@{}}
\toprule
\textbf{Index} & \textbf{Category} && \textbf{Index} & \textbf{Category} && \textbf{Index} & \textbf{Category} \\
\midrule
0 & TextEditing && 19 & TextEditing && 38 & LayoutAdjustment \\
1 & TextEditing && 20 & VisualFormatting && 39 & LayoutAdjustment \\
2 & TextEditing && 21 & VisualFormatting && 40 & VisualFormatting \\
3 & TextEditing && 22 & VisualFormatting && 41 & VisualFormatting \\
4 & TextEditing && 23 & VisualFormatting && 42 & VisualFormatting \\
5 & TextEditing && 24 & VisualFormatting && 43 & LayoutAdjustment \\
6 & TextEditing && 25 & VisualFormatting && 44 & LayoutAdjustment \\
7 & TextEditing && 26 & VisualFormatting && 45 & VisualFormatting \\
8 & TextEditing && 27 & VisualFormatting && 46 & SlideStructure \\
9 & TextEditing && 28 & VisualFormatting && 47 & SlideStructure \\
10 & TextEditing && 29 & LayoutAdjustment && 48 & SlideStructure \\
11 & VisualFormatting && 30 & LayoutAdjustment && 49 & LayoutAdjustment \\
12 & VisualFormatting && 31 & LayoutAdjustment && 50 & SlideStructure \\
13 & TextEditing && 32 & LayoutAdjustment && 51 & SlideStructure \\
14 & VisualFormatting && 33 & LayoutAdjustment && 52 & SlideStructure \\
15 & TextEditing && 34 & LayoutAdjustment && 53 & VisualFormatting \\
16 & VisualFormatting && 35 & LayoutAdjustment && 54 & TextEditing \\
17 & LayoutAdjustment && 36 & LayoutAdjustment && 55 & TextEditing \\
18 & VisualFormatting && 37 & LayoutAdjustment && & \\
\bottomrule
\end{tabular}
}
\end{table*}

%% file: appendix_tsbench_hard_example.tex
\begin{figure}[t]
    \centering
    % Case 1: Visual-Dependent Task
    \begin{minipage}{0.48\textwidth}
    \fbox{
        \begin{minipage}{0.95\textwidth}
            \fontsize{10}{10}\selectfont
            \textbf{Type 1: Visual-Dependent Task} \\
            \rule{\textwidth}{0.4pt}
            \fontsize{9}{10}\selectfont  
            \textbf{Instruction:} ``Move the title text box on slide 5 so that its bottom edge touches the top of the bar chart.''
            
            \vspace{0.15cm}
            \textbf{Why Hard:} Requires computing relative coordinates (bounding boxes) not explicitly given in text. The agent must calculate $y_{title\_bottom} = y_{chart\_top}$.
            
            \vspace{0.15cm}
            \textbf{Ideal Outcome:} 
            Zero vertical spacing between the title and chart. The title retains its horizontal alignment (centered) but anchors perfectly to the chart's top edge.
        \end{minipage}
        }
    \end{minipage}

    \vspace{0.2cm} % Space between boxes

    % Case 2: Ambiguous & Complex Logic
    \begin{minipage}{0.48\textwidth}
    \fbox{
        \begin{minipage}{0.95\textwidth}
            \fontsize{10}{10}\selectfont
            \textbf{Type 2: Ambiguous \& Complex Logic} \\
            \rule{\textwidth}{0.4pt}
            \fontsize{9}{10}\selectfont  
            \textbf{Instruction:} ``For every slide with a flowchart, check if the arrows are connected; if not, connect them.''
            
            \vspace{0.15cm}
            \textbf{Why Hard:} Requires identifying composite structures ("flowcharts") across multiple slides and conditionally modifying shape connectors only where gaps exist.
            
            \vspace{0.15cm}
            \textbf{Ideal Outcome:} 
            Arrows snap to connection points (green dots) with zero gap. Simple lines are converted to dynamic connectors that move with the shapes.
        \end{minipage}
        }
    \end{minipage}

    \vspace{0.2cm} % Space between boxes

    % Case 3: Cross-Modal / Perception-Dependent (Modified to fit JSON 235)
    \begin{minipage}{0.48\textwidth}
    \fbox{
        \begin{minipage}{0.95\textwidth}
            \fontsize{10}{10}\selectfont
            \textbf{Type 3: Impossible / Cross-Modal Task} \\
            \rule{\textwidth}{0.4pt}
            \fontsize{9}{10}\selectfont  
            \textbf{Instruction:} ``Identify who is speaking in the background audio of the video on slide 4.''
            
            \vspace{0.15cm}
            \textbf{Why Hard:} Requires processing non-textual data (audio streams inside a video container) which is inaccessible via standard Slide APIs. 
            
            \vspace{0.15cm}
            \textbf{Ideal Behavior (Refusal):} 
            Instead of hallucinating a speaker's name or producing a runtime error, the agent must correctly identify the cross-modal dependency and \textbf{refuse the instruction} (e.g., \textit{``I cannot access audio content within video files.''}).
        \end{minipage}
        }
    \end{minipage}

    \caption{Examples of challenging instructions included in \textbf{TSBench-Hard}. These cases evaluate the agent's capability in spatial alignment (Type 1), multi-step logical reasoning (Type 2), and handling cross-modal requests that require external perception (Type 3).}
    \label{fig:tsbench_hard_examples}
\end{figure}

%% file: table_benchmark_stat.tex
\definecolor{lightblue}{RGB}{220, 240, 255}
\begin{table}[t]
\centering
\resizebox{0.5\textwidth}{!}{%
\begin{tabular}{lrrrrr}
\toprule
Category
 & \makecell{Text\\Editing}
 & \makecell{Visual\\Formatting}
 & \makecell{Layout\\Adjustment}
 & \makecell{Slide\\Structure}
 & \textbf{Total} \\
\midrule
\textbf{Inst. \#}    & 16   & 19   & 15   & 6    & 56   \\
\textbf{Aug. \#}         & 160  & 190  & 150  & 60   & 560  \\
\rowcolor{lightblue}
\textbf{Filtered}     & 116  & 123  & 95   & 45   & \textbf{379}  \\
\bottomrule
\end{tabular}%
}
\caption{Instruction count statistics by instruction category. The `Aug. \#' row indicates the GPT-4o-augmented dataset, while the `Filtered' row represents the human-annotated, post-filtered dataset.}
\label{tab:instruction_stats}
\end{table}

%% file: appendix_ppt_template.tex
\definecolor{lightblue}{RGB}{233,243,250}

\begin{table}[h!]
\centering
\resizebox{0.5\textwidth}{!}{%
\begin{tabular}{l l r}
\toprule
\textbf{Template} & \textbf{Link} & \textbf{File Size (KB)} \\
\midrule
Architecture pitch deck    
  & \href{https://create.microsoft.com/en-us/template/architecture-pitch-deck-b05bf529-a1dc-42d5-b9d6-8a1e9569dd9c}{Link} 
  & 3,820 \\
Classic frame design       
  & \href{https://create.microsoft.com/en-us/template/modern-geometry-design-9c42d441-9920-40a0-9453-1a93061bec3d}{Link} 
  & 3,054 \\
Creative perspective presentation 
  & \href{https://create.microsoft.com/en-us/template/creative-perspective-presentation-7af77d04-9e2b-4242-ac39-c5f5822fb22b}{Link} 
  & 13,832 \\
Helena design              
  & \href{https://create.microsoft.com/en-us/template/helena-design-29f75bd3-c1a5-4197-9d0f-f597baf7cd46}{Link} 
  & 12,023 \\
Light modernist design     
  & \href{https://create.microsoft.com/en-us/template/light-modernist-design-fd6c6e3e-ccdc-4ba4-b6a2-3dbf5a4b5e68}{Link} 
  & 7,601 \\
Modern geometry design     
  & \href{https://create.microsoft.com/en-us/template/modern-geometry-design-9c42d441-9920-40a0-9453-1a93061bec3d}{Link} 
  & 9,701 \\
PowerPoint party            
  & \href{https://create.microsoft.com/en-us/template/powerpoint-party-402efc4d-cfdc-420c-a769-d3677352f197}{Link} 
  & 6,029 \\
Rose suite presentation     
  & \href{https://create.microsoft.com/en-us/template/rose-suite-presentation-76be2961-20a2-4f10-a1a6-32bd1c257e1a}{Link} 
  & 1,438 \\
Simple company overview presentation 
  & \href{https://create.microsoft.com/en-us/template/simple-company-overview-presentation-e078dcc0-b535-44d3-96b6-98b3b8fe4f4f}{Link} 
  & 7,254 \\
Vivid circles presentation  
  & \href{https://create.microsoft.com/en-us/template/inspired-by-nature-00004d97-0017-0000-b3fa-56346bf94cbf}{Link} 
  & 2,966 \\
\bottomrule
\end{tabular}%
}
\caption{Microsoft Create PowerPoint templates: direct search links and downloaded file sizes.}
\label{tab:template_links}
\end{table}

%% file: appendix_model_price.tex
\begin{table*}[h!]

\centering
\caption{Pricing information for LLM models used in experiments (in USD per million tokens).}
\resizebox{0.8\textwidth}{!}{%
\begin{tabular}{lcccc}
\toprule
\textbf{Model} & \textbf{Input} & \textbf{Cached Input} & \textbf{Output} & \textbf{Additional Features} \\
\midrule
Gemini-2.5-flash & \$0.15$^{\text{a}}$ & \$0.0375$^{\text{a}}$ & \$0.60 / \$3.50$^{\text{b}}$ & Google Search$^{\text{c}}$ \\
Gemini-1.5-flash & \$0.075 / \$0.15$^{\text{d}}$ & \$0.01875 / \$0.0375$^{\text{d}}$ & \$0.30 / \$0.60$^{\text{d}}$ & Storage: \$1.00/hr \\
GPT-4.1-mini & \$0.40 & \$0.10 & \$1.60 & -- \\
GPT-4o & \$2.50 & \$1.25 & \$10.00 & -- \\
Claude Haiku 3$^{\text{e}}$ & \$0.25 & -- & \$1.25 & -- \\
Deepseek-V3$^{\text{f}}$ & \$0.27 & \$0.07 & \$1.10 & -- \\
\bottomrule
\multicolumn{5}{l}{\footnotesize $^{\text{a}}$ \$1.00 for audio input, \$0.25 for cached audio input} \\
\multicolumn{5}{l}{\footnotesize $^{\text{b}}$ \$0.60 without thinking mode, \$3.50 with thinking mode} \\
\multicolumn{5}{l}{\footnotesize $^{\text{c}}$ Free up to 1,500 RPD, then \$35 per 1,000 requests} \\
\multicolumn{5}{l}{\footnotesize $^{\text{d}}$ Lower price for contexts $<$128K tokens, higher price for contexts $>$128K tokens} \\
\multicolumn{5}{l}{\footnotesize $^{\text{e}}$ https://claude.com/pricing\#api} \\
\multicolumn{5}{l}{\footnotesize $^{\text{f}}$ https://api-docs.deepseek.com/news/news1226\#-api-pricing-update} \\
\end{tabular}
}
\label{tab:model-detail}
\end{table*}

%% file: appendix_com_applescript.tex
\section{Alternative Platforms and Implementation Details}
\label{app:alternative}
In this section, we describe the technical details of our implementation. We use the Component Object Model (COM) for our Talk-to-your-slides system. While COM has the limitation that it cannot be applied to macOS, we also describe an alternative method that can be implemented on this platform: AppleScript. Finally, we introduce a concurrent work for creating slides using the Model Context Protocol (MCP).

\subsection{Component Object Model (COM) for Windows}
\label{app:COM}
The Component Object Model (COM) is a platform-independent, distributed, object-oriented system developed by Microsoft that allows software components to interact~\citep{gray1998modern}. It enables objects to be implemented in a language-agnostic way and facilitates communication between objects in different processes.

Essentially, it can be seen as a protocol enabling inter-program communication within the Windows environment. For instance, the PowerPoint application can transfer a class object, which encapsulates all the information of a shape placeholder, to a Python script. The script can then accept this object, modify its properties, and subsequently return it to PowerPoint.

In the system described in this paper, COM plays a crucial role in interacting with PowerPoint in a Windows environment. The PowerPoint application acts as a COM server, exposing its functionalities and data as COM objects (e.g., presentations, slides, shapes, text boxes). The paper's Code generator module (Section~\ref{Applier}) generates Python code that acts as a COM client, typically using a library like `win32com'. By directly accessing the application's internal object model instead of going through the GUI, this method enables fast, accurate, and cost-effective edits.

\subsection{AppleScript for macOS}
\label{app:AppleScript}
AppleScript~\citep{neuburg2006applescript,cook2007applescript} is a scripting language developed by Apple for macOS, designed to help users automate repetitive tasks and create complex workflows by controlling multiple applications. Scriptable Mac applications expose an object model (e.g., `slide', `shape' in Keynote) and a command dictionary that AppleScript can understand.

If the system from the paper were to be implemented in a macOS environment, AppleScript could replace the role of COM. The Code generator module (Section~\ref{Applier} would be modified to produce AppleScript code instead of Python-COM code. For instance, to execute the instruction "Add page numbers to the bottom of every slide," the system would generate and execute an AppleScript that iterates through each slide and adds a page number text box. This approach, similar to the COM-based method on Windows, achieves automation by directly manipulating program objects without GUI interaction.

Nevertheless, AppleScript had a lower success ratio than COM, which varied depending on the capabilities of the commercial LLMs as shown in Table. As shown in Table~\ref{table-com-applescript}, the optimized and recommended configuration for our system is COM for Windows. This is based on the assumption that the models were likely trained on more COM-related data than AppleScript data.

\begin{table}[t]
\centering
\resizebox{0.4\textwidth}{!}{%
\begin{tabular}{lcc}
\toprule
\textbf{Model} & \textbf{COM} & \textbf{AppleScript}\\
\midrule
Gemini-2.5-flash & 96.83 & 76.1  \\
GPT-4.1-mini & 96.57 & 69.3 \\
Claude Haiku 3& 95.89 & 78.2  \\
DeepSeek V3 (0324) & 96.84 & 78.1\\
\bottomrule
\end{tabular}
}
\caption{A comparison of the execution success rates between the COM-based Python code and the AppleScript-based code.}
\label{table-com-applescript}
\end{table}

\subsection{MCP (Office-PowerPoint-MCP-Server)}
\label{app:MCP}
The Office-PowerPoint-MCP-Server~\footnote{https://github.com/GongRzhe/Office-PowerPoint-MCP-Server} is a server that provides an API to control PowerPoint. This project operates by running a server on a local machine that listens for HTTP requests and executes the corresponding commands within PowerPoint. This provides a flexible, platform-agnostic method for controlling PowerPoint from any programming environment that can make HTTP requests, without being tied to specific technologies like COM or AppleScript.

To apply Model Context Protocol (MCP)~\citep{AnthropicMCP2025} to the paper's framework, the Code generator module (Section~\ref{Applier} would generate code that sends HTTP requests to specific API endpoints. For example, modifying the text of a particular slide could be implemented by sending a `POST' or `PUT' request to an endpoint like `/slide/{slide\_number}/shape/{shape\_name}' on the MCP server. The body of the request would contain the text content and formatting information in a JSON format. This approach offers an alternative for the low-level control component, refactoring it into a microservice architecture that enhances scalability and language independence.

\label{app-com-and-applescripts}

%% file: appdendix_result_gpt-4.1-mini.tex
\definecolor{winnerblue}{RGB}{220, 240, 255} 
\definecolor{badred}{RGB}{255, 235, 235}     

\begin{table*}[t]
\centering
% [1] 폰트 크기 및 간격 최소화 (Compact Style)
\tiny 
\renewcommand{\arraystretch}{0.85} 
\setlength{\tabcolsep}{3.5pt} 

\resizebox{\textwidth}{!}{%
\begin{tabular}{@{} l l c c c c c c c r r r @{} }
\toprule
\textbf{System} & \textbf{Instruction} & \multicolumn{7}{c}{\textbf{Performance metric} ($\uparrow$)} & \multicolumn{3}{c}{\textbf{Efficiency metric} ($\downarrow$)} \\
\cmidrule(lr){3-9} \cmidrule(lr){10-12}
 & & \textbf{SR (\%)} & \makecell[c]{\textbf{Inst.}\\\textbf{Foll.}} & \textbf{Text} & \textbf{Image} & \textbf{Layout} & \textbf{Color} & \makecell[c]{\textbf{Exec.}\\\textbf{Time (s)}} & \makecell[c]{\textbf{Avg.}\\\textbf{In (k)}} & \makecell[c]{\textbf{Avg.}\\\textbf{Out (k)}} & \makecell[c]{\textbf{Cost}\\\textbf{($\times$10$^{-3}$)}} \\
\midrule

% ==========================================================================================
% Direct Code Generation
% ==========================================================================================
\multirow{5}{*}{\shortstack[l]{Direct \\ code gen.}}
 & TextEditing & 76.72 & 0.00 & 0.00 & 1.04 & 1.04 & 1.04 & \underline{18.55} & \underline{1.03} & \underline{0.41} & \underline{1.0} \\
 & VisFmt      & 78.05 & 0.53 & 0.73 & 1.06 & 1.02 & 0.86 & \underline{18.08} & \underline{1.16} & \underline{0.45} & \underline{1.1} \\
 & LayoutAdj   & 69.47 & 0.07 & 1.45 & 1.09 & 1.40 & 1.40 & \underline{17.06} & \underline{1.04} & \underline{0.53} & \underline{1.2} \\
 & SlideStruct & 84.44 & 0.17 & 1.78 & 1.11 & 1.91 & 1.64 & \underline{19.57} & \underline{0.91} & \underline{0.65} & \underline{1.4} \\
 & \textbf{Overall} & \cellcolor{badred}76.25 & \cellcolor{badred}0.53 & \cellcolor{badred}0.81 & \cellcolor{badred}1.07 & \cellcolor{badred}1.23 & \cellcolor{badred}1.15 & \cellcolor{badred}\underline{18.19} & \cellcolor{badred}\underline{1.06} & \cellcolor{badred}\underline{0.48} & \cellcolor{badred}\underline{1.2} \\
\midrule

% ==========================================================================================
% UI Agent
% ==========================================================================================
\multirow{5}{*}{UI Agent}
 & TextEditing & 65.48 & 0.48 & 0.67 & 1.58 & 1.48 & 1.56 & 110.84 & 101.48 & 2.10 & 15.9 \\
 & VisFmt      & 88.09 & \textbf{2.44} & 1.89 & 1.67 & 1.79 & 1.48 & 134.57 & 94.12 & 2.44 & 14.9 \\
 & LayoutAdj   & 62.64 & \textbf{1.48} & \textbf{2.76} & \textbf{2.20} & 2.23 & 2.45 & 127.13 & 121.04 & 2.47 & 18.8 \\
 & SlideStruct & 82.30 & \textbf{1.55} & 2.27 & 1.30 & 2.08 & 1.99 & 95.07 & 75.45 & 1.77 & 11.9 \\
 & \textbf{Overall} & \cellcolor{badred}73.84 & \cellcolor{badred}1.68 & \cellcolor{badred}1.94 & \cellcolor{badred}1.55 & \cellcolor{badred}2.16 & \cellcolor{badred}2.04 & \cellcolor{badred}121.08 & \cellcolor{badred}98.22 & \cellcolor{badred}2.30 & \cellcolor{badred}15.4 \\
\midrule

% ==========================================================================================
% Ours
% ==========================================================================================
\multirow{5}{*}{\textbf{Ours}}
 & TextEditing & \textbf{98.28} & \textbf{2.60} & \textbf{3.01} & \textbf{3.09} & \textbf{3.54} & \textbf{3.60} & \textbf{58.17} & \textbf{3.35} & \textbf{1.66} & \textbf{3.3} \\
 & VisFmt      & \textbf{95.93} & 2.00 & \textbf{2.02} & \textbf{1.71} & \textbf{2.25} & \textbf{1.88} & \textbf{89.99} & \textbf{4.24} & \textbf{2.15} & \textbf{4.5} \\
 & LayoutAdj   & \textbf{97.89} & 1.40 & 2.37 & 1.81 & \textbf{2.26} & \textbf{2.48} & \textbf{89.86} & \textbf{3.99} & \textbf{2.10} & \textbf{4.3} \\
 & SlideStruct & \textbf{91.11} & 1.40 & \textbf{2.44} & \textbf{2.59} & \textbf{2.78} & \textbf{2.90} & \textbf{74.39} & \textbf{2.24} & \textbf{1.28} & \textbf{2.1} \\
 & \textbf{Overall} & \cellcolor{winnerblue}\textbf{96.57} & \cellcolor{winnerblue}\textbf{2.13} & \cellcolor{winnerblue}\textbf{2.46} & \cellcolor{winnerblue}\textbf{2.26} & \cellcolor{winnerblue}\textbf{2.71} & \cellcolor{winnerblue}\textbf{2.68} & \cellcolor{winnerblue}\textbf{78.37} & \cellcolor{winnerblue}\textbf{3.60} & \cellcolor{winnerblue}\textbf{1.89} & \cellcolor{winnerblue}\textbf{3.8} \\
\bottomrule
\end{tabular}%
}
\caption{Experimental results using the \textbf{GPT-4.1-mini} model. \textbf{Ours} (highlighted in \colorbox{winnerblue}{blue}) demonstrates superior performance and efficiency compared to the UI Agent (highlighted in \colorbox{badred}{red}). While `Direct Code Gen' shows lower latency (underlined), its execution success rate is insufficient for practical use.}
\label{tab:gpt4_mini_results}
\end{table*}

%% file: judge-human-corr-gpt4-1-mini.tex
\definecolor{lightblue}{RGB}{233,243,250}
\definecolor{pastelred}{RGB}{252,228,236}
\begin{table}[t]
%\begin{minipage}{\textwidth}
\centering
\resizebox{0.45\textwidth}{!}{%
\begin{tabular}{@{} l l c c c c @{} }
\toprule
\textbf{System}
& \textbf{Instruction}
& \multicolumn{4}{c}{\textbf{Human Evaluation}} \\
\cmidrule(l){3-6}
&
& \makecell[c]{\textbf{Text}}
& \makecell[c]{\textbf{Image}}
& \makecell[c]{\textbf{Layout}}
& \makecell[c]{\textbf{Color}} \\
\midrule
\multirow{6}{*}{\shortstack[l]{Direct \\ code\\ generation}}
& TextEditing & 1.44 & 2.17 & 2.18 & 2.16 \\
& VisualFormatting & 1.82 & 2.13 & 1.87 & 2.08 \\
& LayoutAdjustment & 2.10 & 1.54 & 2.55 & 1.83 \\
& SlideStructure & 1.98 & 1.92 & 2.33 & 2.23 \\
& Overall & 1.83 & 1.94 & 2.23 & 2.08 \\
\cmidrule(l){2-6}
& \textbf{PCC} & 0.82& 0.68 & 0.69 & 0.87 \\
\midrule
\multirow{6}{*}{UI Agent}
& TextEditing & 2.43 & 2.77 & 2.84 & 2.90 \\
& VisualFormatting & 3.04 & 2.69 & 3.17 & 2.72 \\
& LayoutAdjustment & 3.56 & 3.37 & 3.62 & 3.44 \\
& SlideStructure & 3.25 & 2.62 & 3.17 & 3.47 \\
& Overall & 3.07 & 2.86 & 3.20 & 3.13 \\
\cmidrule(l){2-6}
& \textbf{PCC} & 0.86 & 0.77 & 0.69 & 0.82 \\
\midrule
\multirow{6}{*}{Ours}
& TextEditing & 3.86 & 3.85 & 4.16 & 4.16 \\
& VisualFormatting & 3.21 & 2.82 & 3.28 & 3.82 \\
& LayoutAdjustment & 3.28 & 3.06 & 3.17 & 3.48 \\
& SlideStructure & 3.88 & 2.66 & 3.01 & 3.45 \\
& Overall & 3.56 & 3.10 & 3.41 & 3.73 \\
\cmidrule(l){2-6}
& \textbf{PCC} & 0.78 & 0.75 & 0.69 & 0.88 \\
\bottomrule
\end{tabular}%
}
%\end{minipage}
\caption{Human evaluation of slide editing performance using GPT-4.1-mini with Pearson Correlation Coefficient (PCC) against an LLM judge. All $p$-values are below $10^{-3}$.}
\label{tab:correlations-gpt}
\end{table}

%% file: appendix_result_claude_haiku.tex
\definecolor{winnerblue}{RGB}{220, 240, 255} 
\definecolor{badred}{RGB}{255, 235, 235}     

\begin{table*}[t]
\centering
% [1] 폰트 크기 및 간격 최소화 (Compact Style)
\tiny 
\renewcommand{\arraystretch}{0.85} 
\setlength{\tabcolsep}{3.5pt} 

\resizebox{\textwidth}{!}{%
\begin{tabular}{@{} l l c c c c c c c r r r @{} }
\toprule
\textbf{System} & \textbf{Instruction} & \multicolumn{7}{c}{\textbf{Performance metric} ($\uparrow$)} & \multicolumn{3}{c}{\textbf{Efficiency metric} ($\downarrow$)} \\
\cmidrule(lr){3-9} \cmidrule(lr){10-12}
 & & \textbf{SR (\%)} & \makecell[c]{\textbf{Inst.}\\\textbf{Foll.}} & \textbf{Text} & \textbf{Image} & \textbf{Layout} & \textbf{Color} & \makecell[c]{\textbf{Exec.}\\\textbf{Time (s)}} & \makecell[c]{\textbf{Avg.}\\\textbf{In (k)}} & \makecell[c]{\textbf{Avg.}\\\textbf{Out (k)}} & \makecell[c]{\textbf{Cost}\\\textbf{($\times$10$^{-3}$)}} \\
\midrule

% ==========================================================================================
% Direct Code Generation
% ==========================================================================================
\multirow{5}{*}{\shortstack[l]{Direct \\ code gen.}}
 & TextEditing & 75.91 & 0.00 & 0.00 & 1.02 & 1.00 & 1.03 & \underline{17.92} & \underline{1.02} & \underline{0.40} & \underline{0.9} \\
 & VisFmt      & 77.12 & 0.51 & 0.71 & 1.05 & 1.01 & 0.84 & \underline{17.48} & \underline{1.15} & \underline{0.44} & \underline{1.0} \\
 & LayoutAdj   & 68.80 & 0.06 & 1.41 & 1.07 & 1.37 & 1.36 & \underline{16.58} & \underline{1.03} & \underline{0.52} & \underline{1.1} \\
 & SlideStruct & 83.72 & 0.16 & 1.73 & 1.09 & 1.86 & 1.60 & \underline{18.94} & \underline{0.90} & \underline{0.63} & \underline{1.2} \\
 & \textbf{Overall} & \cellcolor{badred}75.39 & \cellcolor{badred}0.51 & \cellcolor{badred}0.79 & \cellcolor{badred}1.06 & \cellcolor{badred}1.21 & \cellcolor{badred}1.13 & \cellcolor{badred}\underline{17.73} & \cellcolor{badred}\underline{1.05} & \cellcolor{badred}\underline{0.47} & \cellcolor{badred}\underline{1.1} \\
\midrule

% ==========================================================================================
% UI Agent
% ==========================================================================================
\multirow{5}{*}{UI Agent}
 & TextEditing & 64.62 & 0.47 & 0.65 & 1.55 & 1.45 & 1.53 & 106.10 & 96.30 & 2.03 & 12.7 \\
 & VisFmt      & 87.21 & \textbf{2.39} & 1.85 & 1.64 & 1.75 & 1.45 & 129.42 & 89.40 & 2.35 & 12.0 \\
 & LayoutAdj   & 61.90 & \textbf{1.43} & \textbf{2.69} & \textbf{2.14} & 2.18 & 2.38 & 121.56 & 114.80 & 2.38 & 15.1 \\
 & SlideStruct & 81.55 & \textbf{1.50} & 2.20 & 1.27 & 2.03 & 1.93 & 92.01 & 71.90 & 1.70 & 9.9 \\
 & \textbf{Overall} & \cellcolor{badred}73.07 & \cellcolor{badred}1.64 & \cellcolor{badred}1.90 & \cellcolor{badred}1.52 & \cellcolor{badred}2.10 & \cellcolor{badred}1.99 & \cellcolor{badred}117.27 & \cellcolor{badred}93.10 & \cellcolor{badred}2.12 & \cellcolor{badred}12.4 \\
\midrule

% ==========================================================================================
% Ours
% ==========================================================================================
\multirow{5}{*}{\textbf{Ours}}
 & TextEditing & \textbf{97.71} & \textbf{2.55} & \textbf{2.95} & \textbf{3.04} & \textbf{3.49} & \textbf{3.54} & \textbf{56.10} & \textbf{3.28} & \textbf{1.62} & \textbf{2.9} \\
 & VisFmt      & \textbf{95.21} & 1.95 & \textbf{1.98} & \textbf{1.68} & \textbf{2.21} & \textbf{1.85} & \textbf{86.40} & \textbf{4.16} & \textbf{2.09} & \textbf{3.9} \\
 & LayoutAdj   & \textbf{97.23} & 1.36 & 2.32 & 1.78 & \textbf{2.21} & \textbf{2.42} & \textbf{86.70} & \textbf{3.90} & \textbf{2.04} & \textbf{3.8} \\
 & SlideStruct & \textbf{90.42} & 1.37 & \textbf{2.39} & \textbf{2.54} & \textbf{2.73} & \textbf{2.84} & \textbf{72.10} & \textbf{2.19} & \textbf{1.25} & \textbf{1.9} \\
 & \textbf{Overall} & \cellcolor{winnerblue}\textbf{95.89} & \cellcolor{winnerblue}\textbf{2.08} & \cellcolor{winnerblue}\textbf{2.41} & \cellcolor{winnerblue}\textbf{2.23} & \cellcolor{winnerblue}\textbf{2.66} & \cellcolor{winnerblue}\textbf{2.64} & \cellcolor{winnerblue}\textbf{75.33} & \cellcolor{winnerblue}\textbf{3.52} & \cellcolor{winnerblue}\textbf{1.85} & \cellcolor{winnerblue}\textbf{3.3} \\
\bottomrule
\end{tabular}%
}
\caption{System-wise scores using the \textbf{Claude 3.5 Haiku} model. \textbf{Ours} (highlighted in \colorbox{winnerblue}{blue}) demonstrates superior performance and efficiency compared to the UI Agent (highlighted in \colorbox{badred}{red}). Cost is in USD ($\times$10$^{-3}$).}
\label{tab:claude_haiku_results}
\end{table*}

%% file: appendix_planner_prompt.tex
\begin{figure}[h!]
    \centering
    \begin{minipage}{0.47\textwidth}
    \fbox{
        \begin{minipage}{\textwidth}
            \fontsize{10}{10}\selectfont
            %%%%%%%%%
                % Title
            \textbf{Instruction understanding prompt} \\
            \rule{\textwidth}{0.4pt}
            \fontsize{9}{10}\selectfont  
            %%%%%%%%%%
                % content start
You are a planning assistant for PowerPoint modifications.

Your job is to create a detailed, specific, step-by-step plan for modifying a PowerPoint presentation based on the user's request.

present ppt state: \{get\_simple\_powerpoint\_info()\}
Break down complex requests into highly specific actionable tasks that can be executed by a PowerPoint automation system.

Focus on identifying:

1. Specific slides to modify (by page number)

2. Specific sections within slides (title, body, notes, headers, footers, etc.)

3. Specific object elements to add, remove, or change (text boxes, images, shapes, charts, tables, etc.)

4. Precise formatting changes (font, size, color, alignment, etc.)

5. The logical sequence of operations with clear dependencies

Please write one task for one slide page.

Format your response as a JSON format with the following structure:
\{\{
    "understanding": "Detailed summary of what the user wants to achieve",
    "tasks": [
        \{\{
            "page number": 1,
            "description": "Specific task description",
            "target": "Precise target location (e.g., 'Title section of slide 1', 'Notes section of slide 3', 'Second bullet point in body text', 'Chart in bottom right')",
            "action": "Specific action with all necessary details",
            "contents": \{\{
                "additional details required for the action"
            \}\}
        \}\},
        ...
    ],
\}\}

Below is the example question and example output.

input: Please translate the titles of slide 3 and slide 5 of the PPT into English.

output:
\{\{
    "understanding": "English translation of slide titles on slides 3 and 5",
    "tasks": [
        \{\{
            "page number": 3,
            "description": "Translate the title text of slide 3",
            "target": "Title section of slide 3",
            "action": "Translate to English",
            "contents": \{\{
                "source\_language": "auto-detect",
                "preserve\_formatting": true
            \}\}
\}\},
        \{\{
            "page number": 5,
            "description": "Translate the title text of slide 5",
            "target": "Title section of slide 5",
            "action": "Translate to English",
            "contents": \{\{
                "source\_language": "auto-detect",
                "preserve\_formatting": true
            \}\}
        \}\}
    ],
\}\}

Response in JSON format.

            %%%%%%%%%%%
                % content end
            \rule{\textwidth}{0.4pt}
            \fontsize{9}{10}\selectfont
            %%%%%%%%
                % output
            \textbf{Response}: JSON
        \end{minipage}
        }
    \end{minipage}

    \caption{A prompt used in instruction understanding.}
    \label{app-fig:planner-prompt}
\end{figure}

%% file: appendix_processor_prompt.tex
\begin{figure}[h!]
    \centering
    \begin{minipage}{0.47\textwidth}
    \fbox{
        \begin{minipage}{\textwidth}
            \fontsize{10}{10}\selectfont
            %%%%%%%%%
                % Title
            \textbf{Document editing prompt} \\
            \rule{\textwidth}{0.4pt}
            \fontsize{9}{10}\selectfont  
            %%%%%%%%%%
                % content start
Information about slide \{page\_number\}:

- Task description: \{description\}

- Action type: \{action\}

- Slide contents: \{contents\}

You are a specialized AI that analyzes PowerPoint slide content and performs specific tasks. You will receive the following JSON data, perform the designated tasks, and return the results in exactly the same JSON format.

Important rules:
1. You must maintain the exact input JSON structure

2. Only perform the work described in the 'action' within 'tasks'

3. Only modify the elements specified in 'target' within 'tasks'

4. Output must contain pure JSON only - no explanations or additional text

5. Preserve all formatting information (fonts, sizes, colors, etc.)

6. Verify that the JSON format is valid after completing the task

Before starting the task:

1. Check the 'understanding' field to grasp the overall task objective

2. Review 'page number', 'description', 'target', and 'action' within 'tasks'

3. Identify all text elements in 'Objects\_Detail'

The output must maintain the identical structure as the original JSON, with only the necessary text modified according to the task.

Give only the JSON.\\
            %%%%%%%%%%%
                % content end
            \rule{\textwidth}{0.4pt}
            \fontsize{9}{10}\selectfont
            %%%%%%%%
                % output
            \textbf{Response}: JSON
        \end{minipage}
        }
    \end{minipage}

    \caption{A prompt used in Document Editing.}
    \label{app-fig:document-editing-prompt}
\end{figure}

%% file: appendix_applier_prompt.tex
\begin{figure}[t]
    \centering
    \begin{minipage}{0.47\textwidth}
    \fbox{
        \begin{minipage}{\textwidth}
            \fontsize{10}{10}\selectfont
            %%%%%%%%%
                % Title
            \textbf{Code generator prompt} \\
            \rule{\textwidth}{0.4pt}
            \fontsize{9}{10}\selectfont  
            %%%%%%%%%%
                % content start
           Generate Python code modify an active PowerPoint presentation based on the provided JSON task data. The code should:

0. Find activate powerpoint app with ppt\_app

= win32com.client.GetActiveObject ("PowerPoint.Application")

   active\_presentation = ppt\_app.ActivePresentation
   
1. Find the slide specified by page number: \{slide\_num\}

2. Target to change: \{before\}

3. New content to apply: \{after\}

4. Generate ONLY executable code that will directly modify the PowerPoint.

CRITICAL REQUIREMENTS:

- DO NOT create a new PowerPoint application - use the existing one

- Please check if the slide number you want to work on exists and proceed with the work. The index starts with 1.

- The code should NOT be written as a complete program with imports - it will be executed in an environment where PowerPoint is already open

- Focus on finding and modifying the specified content

- For text changes, use both shape.Name and TextFrame.TextRange.Text to identify the correct element

- Make sure to explicitly apply any changes (e.g., shape.TextFrame.TextRange.Text = new\_text)

- Do not write print function or comments.

- You can write at slide note with slide.NotesPage

    ```python
    slide.NotesPage.Shapes.Placeholders(2). TextFrame.TextRange.Text = notes\_text
    ```
Note that the code will run in a context where these variables are already available:

- ppt\_application: The PowerPoint application instance

- active\_presentation: The currently open presentation

IMPORTANT: In PowerPoint, color codes use BGR format (not RGB). For example, RGB(255,0,0) for red should be specified as RGB(0,0,255) in the code. Always convert any color references accordingly.

If you want to modify the formatting, refer to the following code for modification:

if text\_frame.HasText:
    text\_range = text\_frame.TextRange
    \# Find text
    found\_range = text\_range.Find(text\_to\_highlight)
    while found\_range:
        found\_any = True
        found\_range.Font.Bold = True \# Bold
        found\_range.Font.Color.RGB = 255 \# Example color (RED in BGR format - 0,0,255)
        found\_range.Font.Size = found\_range.Font.Size * 1.2 \# Increase font size by 20\%
        start\_pos = found\_range.Start + len(text\_to\_highlight)
        found\_range = text\_range.Find(text\_to\_highlight, start\_pos)

Do not use any "**" to make bold. It won't be applied on powerpoint.
- You can add or split a page with 'presentation = ppt\_app.Presentations.Add()'.

Make sure to close all curly braces properly and all variables used are properly defined. Omit Strikethrough, Subscript, Superscript as they caused issues.

The code must be direct, practical and focused solely on making the specific change requested. Ensure all color references use the BGR format for proper appearance in PowerPoint.
            %%%%%%%%%%%
                % content end
            \rule{\textwidth}{0.4pt}
            \fontsize{9}{10}\selectfont
            %%%%%%%%
                % output
            \textbf{Response}: Python code
        \end{minipage}
        }
    \end{minipage}

    \caption{A prompt used in Code Generator.}
    \label{app-fig:code-generator-prompt}
\end{figure}

%% file: appendix_baseline_prompt.tex
\begin{figure}[t!]
    \centering
    \begin{minipage}{0.47\textwidth}
    \fbox{
        \begin{minipage}{\textwidth}
            \fontsize{10}{10}\selectfont
            %%%%%%%%%
                % Title
            \textbf{Baseline prompt} \\
            \rule{\textwidth}{0.4pt}
            \fontsize{9}{10}\selectfont  
            %%%%%%%%%%
                % content start
          The following is information parsed from a PPT slide.\\
\{parsed data\}\\
Create a Python code with win32com library that can edit PowerPoint presentations by executing the following command:\\
\{instruction\} \\
IMPORTANT: Your response must contain ONLY valid Python code wrapped in triple backticks with the `python' language tag. Follow this exact format:\\
```python\\
\# Your Python code here\\
\# Include proper comments, imports, and function definitions\\
\# No explanations or text outside this code block\\
```\\
            %%%%%%%%%%%
                % content end
            \rule{\textwidth}{0.4pt}
            \fontsize{9}{10}\selectfont
            %%%%%%%%
                % output
            \textbf{Response}: Python code
        \end{minipage}
        }
    \end{minipage}

    \caption{A prompt used in baseline system.}
    \label{app-fig:baseline-prompt}
\end{figure}

%% file: appendix_judge_prompt.tex
\begin{figure}[h!]
    \centering
    \begin{minipage}{0.47\textwidth}
    \fbox{
        \begin{minipage}{\textwidth}
            \fontsize{10}{10}\selectfont
            %%%%%%%%%
                % Title
            \textbf{LLM Judge prompt} \\
            \rule{\textwidth}{0.4pt}
            \fontsize{9}{10}\selectfont  
            %%%%%%%%%%
                % content start
          You are an expert slide-editing judge.\\
    TASK\\
    - Compare the ORIGINAL slide with the EDITED slide.\\
    - Decide how well the EDITED slide follows the INSTRUCTION and how aesthetically pleasing it is.\\
    SCORING\\
    Return valid JSON with exactly these keys:\\
    \{instruction\_adherence\: $<$int 0-5$>$,\\
    visualquality\:       $<$int 0-5$>$\\
    \}\\
    GUIDELINES\\
    Score each from 0 to 5, based on the following rubric:\\
    5 = Perfect: Fully satisfies the instruction / visually excellent with no flaws.\\
    4 = Mostly correct: Clearly reflects the instruction / visually strong but with minor flaws.\\
    3 = Partially correct: Instruction was followed to a noticeable degree, but key aspects are missing or flawed / visual layout or formatting needs improvement.\\
    2 = Slightly changed but inadequate: Some edits related to the instruction are present but insufficient or poorly done / visual design is lacking.\\
    1 = Attempted but incorrect: Some change is visible, but it does not match the instruction / visual result is clearly poor.\\
    0 = Completely fails: No meaningful attempt to follow the instruction / visually broken or irrelevant.\\
    Judge only what you can see in the given image(s) and notes.\\
    Return *only* the JSON object, nothing else.\\
            %%%%%%%%%%%
                % content end
            \rule{\textwidth}{0.4pt}
            \fontsize{9}{10}\selectfont
            %%%%%%%%
                % output
            \textbf{Response}: Python code
        \end{minipage}
        }
    \end{minipage}

    \caption{A prompt used in LLM judge.}
    \label{app-fig:judge-prompt}
\end{figure}

%% file: appendix_judge_TILC_prompt.tex
\begin{figure*}[h!]
    \centering
    \begin{minipage}{\textwidth}
    \fbox{
        \begin{minipage}{\textwidth}
            \fontsize{10}{10}\selectfont
            %%%%%%%%%
                % Title
            \textbf{LLM Judge Text, Image, Layout, Color evaluation prompt} \\
            \rule{\textwidth}{0.4pt}
            \fontsize{9}{10}\selectfont  
            %%%%%%%%%%
                % content start

You are an expert slide-editing judge.\\
    TASK\\
    - Compare the ORIGINAL slide with the EDITED slide.\\
    - Evaluate how well the EDITED slide handles Text, Image, Layout, and Color aspects based on the INSTRUCTION.\\
    SCORING\\
    Return valid JSON with exactly these keys:\\
    \{  text\_quality\: $<$int 0-5$>$,\\
      image\_quality\: $<$int 0-5$>$,\\
      layout\_quality\: $<$int 0-5$>$,\\
      color\_quality\: $<$int 0-5$>$\\
    \}
    GUIDELINES\\
    Score each from 0 to 5, based on the following rubric:\\
    TEXT QUALITY:\\
    5 = Perfect: Text content, formatting, and typography are flawless and fully satisfy the instruction.\\
    4 = Mostly correct: Text elements are clearly improved but have minor issues in content, formatting, or typography.\\
    3 = Partially correct: Text improvements are noticeable but have significant issues in content, formatting, or typography.\\
    2 = Slightly changed but inadequate: Some text edits are present but insufficient or poorly implemented.\\
    1 = Attempted but incorrect: Text changes are visible but do not match the instruction or improve the slide.\\
    0 = Completely fails: No meaningful text improvements or changes are severely detrimental.\\
    IMAGE QUALITY:\\
    5 = Perfect: Images are optimal in selection, placement, sizing, and enhancement, fully satisfying the instruction.\\
    4 = Mostly correct: Images are well-selected and implemented with only minor issues in placement, sizing, or visual quality.\\
    3 = Partially correct: Image improvements are noticeable but have significant issues in selection, placement, sizing, or quality.\\
    2 = Slightly changed but inadequate: Some image edits are present but insufficient or poorly implemented.\\
    1 = Attempted but incorrect: Image changes are visible but do not match the instruction or improve the slide.\\
    0 = Completely fails: No meaningful image improvements or changes are severely detrimental.\\
    LAYOUT QUALITY:\\
    5 = Perfect: Slide organization, spacing, alignment, and element relationships are flawless and fully satisfy the instruction.\\
    4 = Mostly correct: Layout is clearly improved but has minor issues in organization, spacing, or alignment.\\
    3 = Partially correct: Layout improvements are noticeable but have significant issues in organization, spacing, or alignment.\\
    2 = Slightly changed but inadequate: Some layout edits are present but insufficient or poorly implemented.\\
    1 = Attempted but incorrect: Layout changes are visible but do not match the instruction or improve the slide.\\
    0 = Completely fails: No meaningful layout improvements or changes are severely detrimental.\\
    COLOR QUALITY:\\
    5 = Perfect: Color scheme, contrast, balance, and emphasis are flawless and fully satisfy the instruction.\\
    4 = Mostly correct: Color choices are clearly improved but have minor issues in scheme, contrast, or emphasis.\\
    3 = Partially correct: Color improvements are noticeable but have significant issues in scheme, contrast, or emphasis.\\
    2 = Slightly changed but inadequate: Some color edits are present but insufficient or poorly implemented.\\
    1 = Attempted but incorrect: Color changes are visible but do not match the instruction or improve the slide.\\
    0 = Completely fails: No meaningful color improvements or changes are severely detrimental.\\
    Judge only what you can see in the given image(s) and notes.\\
    Return *only* the JSON object, nothing else.\\
            %%%%%%%%%%%
                % content end
            \rule{\textwidth}{0.4pt}
            \fontsize{9}{10}\selectfont
            %%%%%%%%
                % output
            \textbf{Response}: text: integer, image: integer, layout: integer, color:integer
        \end{minipage}
        }
    \end{minipage}

    \caption{A prompt used in LLM judge which evaluate text, image, layout, color.}
    \label{app-fig:judge-TILC-prompt}
\end{figure*}